\documentclass{article}

\PassOptionsToPackage{numbers, compress, sort}{natbib}

\usepackage[final]{neurips_2022}

\usepackage[utf8]{inputenc} 
\usepackage[T1]{fontenc}    
\usepackage{hyperref}       
\usepackage{url}            
\usepackage{booktabs}       
\usepackage{amsfonts}       
\usepackage{nicefrac}       
\usepackage{microtype}      
\usepackage{xcolor}         

\usepackage{microtype}
\usepackage{graphicx}
\usepackage{subfigure}
\usepackage{booktabs} 
\usepackage{url}
\usepackage{graphicx}
\usepackage{amsmath}
\usepackage{amssymb}
\usepackage{amsthm}
\usepackage{tikz}
\usepackage{mathrsfs}
\usepackage{bm}
\usepackage{verbatim}
\usepackage{setspace}
\usepackage{xcolor}
\usepackage{enumitem}
\usepackage[super]{nth}
\usepackage[utf8]{inputenc}
\usepackage{multirow}
\usepackage{hhline}
\usepackage{colortbl}
\usepackage{diagbox}
\usepackage{makecell}
\usepackage{tabularx}

\usetikzlibrary{arrows}
\usetikzlibrary{arrows.meta}
\usetikzlibrary{calc}
\usetikzlibrary{decorations}
\usetikzlibrary{decorations.markings}
\usetikzlibrary{external}
\usetikzlibrary{patterns}
\usetikzlibrary{positioning}
\usetikzlibrary{shapes}
\usetikzlibrary{tikzmark}
\usetikzlibrary{fit}

\title{Tree ensemble kernels for Bayesian optimization with known constraints over 
mixed-feature spaces}

\author{%
  Alexander Thebelt \thanks{Corresponding author: alexander.thebelt18@imperial.ac.uk} \\
  Imperial College London\\
  London, UK \\
  \And
  Calvin Tsay \\
  Imperial College London \\
  London, UK \\
  \And
  Robert M. Lee \\
  BASF SE \\
  Ludwigshafen, Germany \\
  \And
  Nathan Sudermann-Merx \\
  Cooperative State University \\
  Mannheim, Germany \\
  \And
  David Walz \\
  BASF SE \\
  Ludwigshafen, Germany \\
  \And
  Behrang Shafei \\
  BASF SE \\
  Ludwigshafen, Germany \\
  \And
  Ruth Misener \\
  Imperial College London \\
  London, UK \\
}

\begin{document}

\maketitle

\begin{abstract}
Tree ensembles can be well-suited for black-box optimization tasks such as algorithm tuning and neural architecture 
    search, as they achieve good predictive performance with little or no manual tuning, naturally handle discrete 
    feature spaces, and are relatively insensitive to outliers in the training data.
Two well-known challenges in using tree ensembles for black-box optimization are (i) effectively 
    quantifying model uncertainty for exploration and (ii) optimizing over the piece-wise
    constant acquisition function.
To address both points simultaneously, we propose using the kernel interpretation of tree ensembles as a
    Gaussian Process prior to obtain model variance estimates, and we develop a compatible
    optimization formulation for the acquisition function.
The latter further allows us to seamlessly integrate known constraints to improve sampling 
    efficiency by considering domain-knowledge in engineering settings and modeling 
    search space symmetries, e.g.,\ hierarchical relationships in neural architecture search.
Our framework performs as well as state-of-the-art methods for unconstrained black-box optimization over continuous/discrete features and outperforms competing methods for problems combining mixed-variable
    feature spaces and known input constraints.
    
\end{abstract}

\section{Introduction} \label{sec:motivation}
Many black-box optimization problems contain feature relationships known \textit{a priori} based on domain knowledge, such as hierarchies or constraints~\citep{nanfack2022constraint, thebelt2022maximizing}. 
For example, hierarchical structures arise in neural architecture search \citep{elsken2019neural,ying2019bench}, where hyperparameters such as kernel size are only relevant if a convolutional layer is selected. 
Explicit constraints can also arise, e.g.,\ matching kernel size and stride to input channel size and padding. 
In many cases, Bayesian optimization can incorporate known hierarchies and/or constraints, given that a suitable surrogate model is selected. 
To this end, \citet{fromont2006integrating} and \citet{nijssen2007mining} impose a variable hierarchy by constraining the splitting order of decision trees, i.e.,\ certain attributes must be selected before others.

Tree-based models, such as random forests or gradient-boosted trees, remain popular in many applications, as they inherit the innate ability of simple decision trees to seamlessly handle categorical and discrete input spaces. 
Moreover, they are highly parallelizable and scalable to high-dimensional data. 
Despite these modelling advantages, the deployment of tree-based models in Bayesian optimization has been limited by challenges pertaining to (i) quantifying prediction uncertainty and (ii) optimizing acquisition functions defined by their discontinuous response surfaces \citep{shahriari2016BO}. 
Early works, e.g.,\ the popular SMAC algorithm \citep{hutter2011SequentialModel}, addressed (i) using empirical variance within a tree ensemble and (ii) via local and/or random search methods.
Moreover, recent works~\citep{misic2017OptimizationEnsembles,mistry2018MixedIntegerEmbedded,thebelt2021entmoot} propose mixed-integer formulations for tree ensembles, enabling optimization over their mean functions. 


\textbf{Contributions.} Sections~\ref{sec:related_work} and \ref{sec:tree_kernel_inference} present related work and methods used to derive the approach proposed in this paper. We present a mixed-integer second-order cone optimization formulation for tree kernel Gaussian processes in Section~\ref{sec:constrained_bayesian_optimization} and 
show that that the \textit{tree agreement ratio}, i.e.,\ the hyperparameter introduced by the tree ensemble kernel, sufficiently represents the model uncertainty in Section~\ref{sec:uncertainty_metric}. 
Section~\ref{sec:local_vs_global} shows that solving the mixed-integer second-order cone optimization problem considerably outperforms sampling-based strategies in Bayesian optimization. Our approach of using tree ensemble kernels as a Gaussian process prior is particularly useful for applications combining mixed-variable spaces and known input constraints. A Python implementation of the proposed algorithm is available at: \url{www.github.com/cog-imperial/tree_kernel_gp}

\section{Related work} \label{sec:related_work}
Bayesian optimization (BO) solves \citep{frazier2018Tutorial, shahriari2016BO, greenhill2020bayesian}:
    $\mathbf{x}_f^* \in \text{arg max}_{\mathbf{x}} \; f(\mathbf{x}),$
where $f$ is an expensive-to-evaluate black-box function that can be queried at inputs $\mathbf{x} \in \mathcal{X}$ to derive the optimal solution $\mathbf{x}_f^*$.
BO iteratively updates a surrogate model of $f$ and optimizes a corresponding acquisition function that balances exploitation and exploration. 
Maximizing the acquisition function produces a new query $\mathbf{x}^*$ which is evaluated and added to the set of observations.
Gaussian processes (GPs) \citep{rasmussen2006GP} are a common choice of BO surrogate  due to their flexibility, e.g.,\ domain-specific knowledge can be built into the GP prior via mean and kernel functions, and reliable uncertainty quantification to identify unexplored search areas.
Open-source tools such as BoTorch \citep{balandat2020botorch} implement BO with GP surrogates and offer a wide selection of kernels mainly suited for continuous search spaces.
Prior works developed GPs with modified kernels to integrate discrete features \citep{ru2020bayesian, garrido2020dealing, deshwal2021bayesian, hase2021gryffin, buathong2020kernels} or considered conditional feature spaces \citep{levesque2017bayesian, jenatton2017bayesian, ma2020additive, han2021high}.
\citet{nguyen2020bayesian} integrate catgorical and category-specific continuous inputs by formulating the black-box optimization problem as a multi-arm bandit problem for which each category corresponds to an arm.
Similarly, \citet{gopakumar2018algorithmic} handle mixed-type inputs using multi-armed bandits.
Besides GPs, tree ensemble-based surrogates show excellent performance for black-box optimization with mixed-variable settings, i.e.,\ with continuous, integer and categorical variables, and for structured search spaces, e.g.,\ hierarchical and conditional feature spaces \citep{hutter2011SequentialModel}.

Black-box optimization tools using tree ensembles, e.g.,\ \texttt{SMAC} \citep{hutter2011SequentialModel} and Scikit-Optimize (\texttt{SKOPT}) \citep{head2018ScikitOptimize}, are useful for applications such as neural architecture search (NAS) and algorithm tuning.
\citet{shahriari2016BO} mention challenges in deploying tree ensembles for BO: (i) quantifying uncertainty for exploration purposes, and (ii) optimizing over the non-differentiable discrete acquisition function to determine the next query point.
\texttt{SMAC} identifies uncertain search space regions using empirical variables across tree predictions of the random forest and optimizes the acquisition function combining local and random search.
\citet{bergstra2011algorithms} proposes the Tree Parzen Estimator (TPE) to handle categorical variables and conditional structures by modeling individual input dimensions by a kernel density estimator. 
However, the TPE approach ignores dependencies between dimensions.
For gradient-boosted tree ensembles, \texttt{SKOPT} derives uncertainty with quantile regression to fit two models for the 16th and 84th percentile and averages the predictions to estimate the standard deviation. 
For random forests, \texttt{SKOPT} uses an uncertainty strategy similar to \texttt{SMAC}.
In general, \texttt{SKOPT} relies on random sampling to optimize the acquisition function.

\citet{misic2017OptimizationEnsembles} proposed a mixed-integer optimization formulation for 
tree ensemble mean functions that has been used in several applications \citep{ceccon2022omlt,mistry2018MixedIntegerEmbedded, thebelt2021entmoot,thebelt2020global, thebelt2022multi}.
Besides improving the solution to an acquisition function, mixed-integer formulations also allow  explicit consideration of input constraints to incorporate domain knowledge.
Some software tools, e.g.,\ BoTorch, support linear equality and inequality constraints of continuous variables at the acquisition function optimization step, while tree ensemble-based algorithms do not support input constraints.
\citet{papalexopoulos2021constrained} use ReLU neural networks as surrogate models and deploy a mixed-integer linear formulation to optimize the acquisition function.
The approach relies on random initialization and stochasticity in the model training to allow for exploration.
\citet{daxberger2019mixed} handle mixed-variable search spaces by using a Bayesian linear regressor that uses an integer solver to search the discrete subspace.
The authors introduce features capturing the discrete parts of the search space by using a BOCS model \citep{baptista2018bayesian, deshwal2020scalable_lit_rev}, while continuous parts are handled with random Fourier features \citep{rahimi2007random}.
Genetic Algorithms (\texttt{GA}) are another class of algorithms, which deploy evolution-based selection heuristics to maximize black-box functions \citep{jin2005Evolutionary}.
While there is no feasibility guarantee for input constraints, \texttt{GA} implementations like pymoo \citep{pymoo} support constraint optimization by minimizing constraint violation. 

We compare our BO approach, which uses the kernel interpretation of tree ensembles as a Gaussian process prior, to: \texttt{SMAC}, the random forest and gradient-boosted tree versions of \texttt{SKOPT}, (\texttt{SKOPT-RF}, and \texttt{SKOPT-GBRT}, respectively) in the Section~\ref{sec:numerical_studies} numerical studies as a baseline for other tree ensemble-based algorithms. 
We also compare against the default upper-confidence bound and expected improvement BO implementations of BoTorch (\texttt{UCB-MATERN} and \texttt{EI-MATERN}, respectively) and the default \texttt{GA} algorithm of pymoo to include black-box algorithms that partially support constrained optimization.

\section{Technical background on prior work} \label{sec:tree_kernel_inference}

\subsection{Tree ensemble kernel as a Gaussian process prior} \label{sec:tree_ensemble_kernel}
Exploring the search space requires quantifying the uncertainty of the underlying surrogate model. 
We use the kernel interpretation of tree ensembles based on random partitions \cite{davies2014random, zafari2019evaluating}. The tree kernel captures correlation between two input data points $(\mathbf{x},\mathbf{x'}) \in \mathbb{R}^n$:
\begin{equation}
    \label{eq:tree_kernel}
    k_{\text{Tree}}(\mathbf{x}, \mathbf{x'}) = 
        \sigma_0^2 \; |\mathcal{T}|^{-1} \;
        \mathbf{z}(\mathbf{x})^\intercal \mathbf{z}(\mathbf{x'})
\end{equation}
To derive the tree kernel, we first train a gradient boosted tree ensemble $\mathcal{T}$ on data set $\mathbf{X} \in \mathbb{R}^{m \times n}$ with $n$ denoting the dimensionality of the search space and $m$ the size of the data set.
Every tree $t$ in the tree ensemble $\mathcal{T}$ maps inputs $\mathbf{x}\in \mathbb{R}^n$ onto a leaf $l$ by sequentially evaluating splitting conditions.
Each leaf $l$ defines a subspace $\mathbf{x}_l \subset \mathbb{R}^n$ restricted by active splits $s \in \mathbf{splits}(t)$.
Two inputs $(\mathbf{x},\mathbf{x'})$ are fully correlated in tree $t$ if both end up in the leaf subspace $\mathbf{x}_{t,l}$ and uncorrelated otherwise.
The Eq.~\eqref{eq:tree_kernel} vector $\mathbf{z}(\mathbf{x})$ consists of binary elements $z_{t,l}$ indicating if leaf $l \in \mathcal{L}_t$ is active for input $\mathbf{x}$, with $\mathcal{L}_t$ denoting the set of all leaves in tree $t$.
The inner product $\mathbf{z}(\mathbf{x})^\intercal \mathbf{z}(\mathbf{x'})$, normalized by the total number of trees $|\mathcal{T}|$, gives the ratio of trees in the ensemble for which $\mathbf{x}$ and $\mathbf{x'}$ fall into the same leaf.
We modify the kernel by adding a trainable signal variance $\sigma_0^2$ \citep{lee2015face}.
Note that the tree ensemble defining the kernel and the kernel hyperparameters are trained separately.
\citet{davies2014random} prove that the tree kernel is a suitable GP prior.
The resulting (non-stationary and supervised) tree kernel describes a prior over piece-wise constant functions when used in a GP.

\subsection{Posterior distribution} \label{sec:posterior_distribution}
We approximate $f$ as a Gaussian process with zero mean and kernel $k_\text{Tree}$:
$
    f(\cdot) \sim \mathcal{GP}(0, k_{\text{Tree}})
$.
Since $k_\text{Tree}$ is a valid Mercer kernel, the mean $M(\mathbf{x})$ and variance $V(\mathbf{x})$ of the GP at 
    $x \in \mathbb{R}^n$ is \cite{rasmussen2006GP}:
\begin{subequations}
    \label{eq:gp_infer}
    \begin{align}
        M(\mathbf{x}) &= K_{\mathbf{x},\mathbf{X}} \; (K_{\mathbf{X},
        \mathbf{X}})^{-1} \; \mathbf{y} \label{eq:a_mean} \\
        V(\mathbf{x}) &= K_{\mathbf{x},\mathbf{x}} - K_{\mathbf{x},\mathbf{X}} \;
        (K_{\mathbf{X},\mathbf{X}})^{-1} \; K_{\mathbf{x},
        \mathbf{X}}^{\intercal} \label{eq:b_var}
    \end{align}
\end{subequations}
The Gram matrix $K_{\mathbf{X},\mathbf{X}} \in \mathbb{R}^{m \times m}$
    has entries describing pairwise correlations computed
    based on the kernel function in Eq.~\eqref{eq:tree_kernel} .
The entries of vector $K_{\mathbf{x},\mathbf{X}} \in \mathbb{R}^{m}$ contain correlations
    between the input $\mathbf{x}$ and sampled data points, defined as
    $\left[ k_\text{Tree}(\mathbf{x}, \mathbf{x}_1), k_\text{Tree}(\mathbf{x}, \mathbf{x}_2), 
    \dotsc, k_\text{Tree}(\mathbf{x}, \mathbf{x}_m)\right]$ with vectors $\mathbf{x}_i$ 
    referring to rows in data set $\mathbf{X}$.
Target values $\mathbf{y} \in \mathbb{R}^m$ are the corresponding observations for $\mathbf{X}$. 
Eq.~\eqref{eq:gp_infer} describes the noise-free case of the GP mean and variance. 
To fit a GP based on the tree kernel function, we usually require a noise term, i.e.,\ a diagonal 
    matrix $\sigma_y^2 I$ that is added to $K_{\mathbf{X},\mathbf{X}}$.
We set hyperparameters $\sigma_y^2$ and $\sigma_0^2$ by maximizing the log marginal likelihood.
\citet{lee2015face} compute the Eq.~\eqref{eq:a_mean} inverse of the Gram matrix $K_{\mathbf{X},\mathbf{X}}$ efficiently by exploiting the property that the rank of $K_{\mathbf{X},\mathbf{X}}$ is at most the number of leaves over trees.
\begin{figure}
    \centering
    \subfigure[$M(\mathbf{x})$]{\label{fig:treed_gp_branin_mean}
        \includegraphics[width=0.323\linewidth]{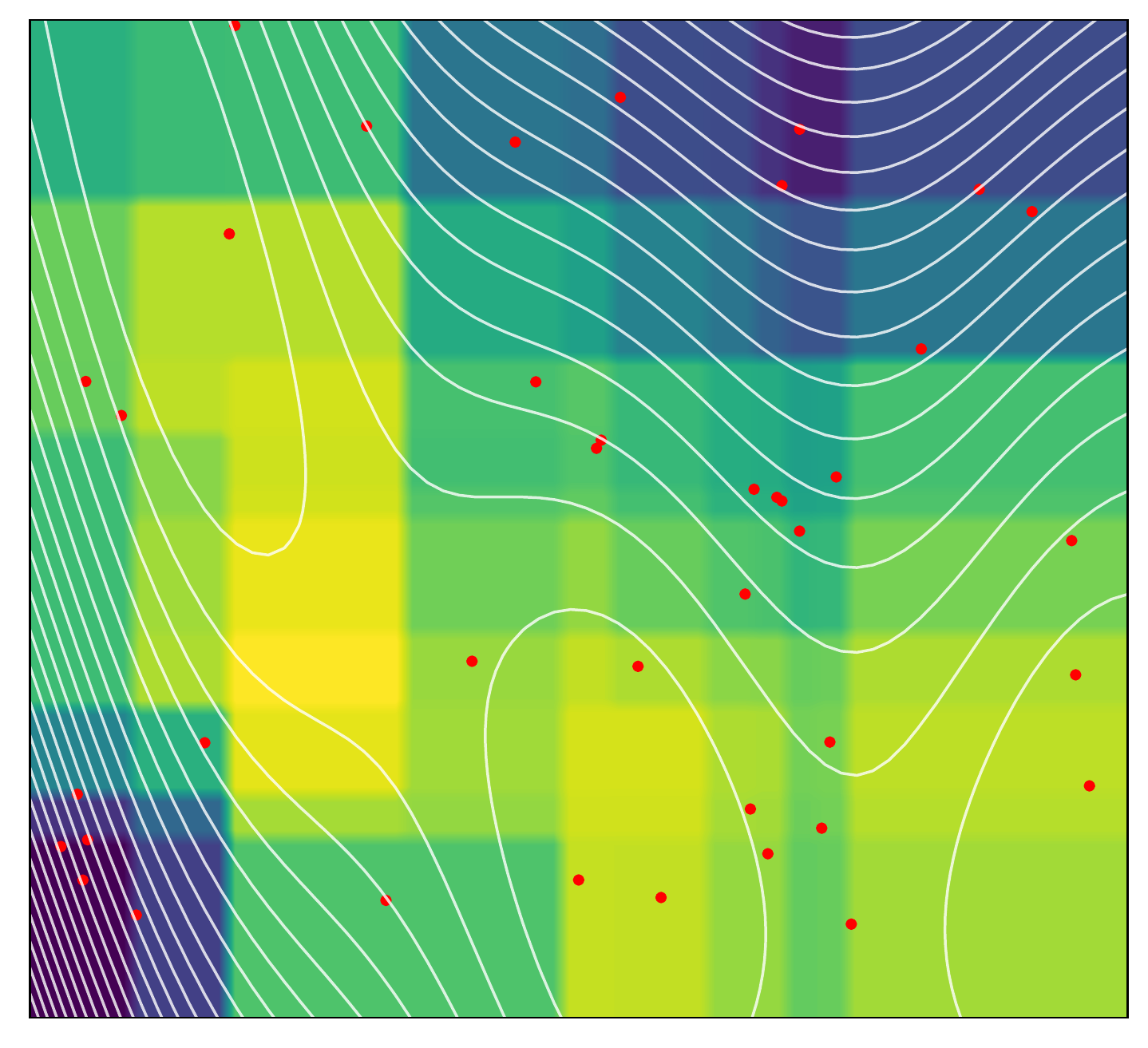}}
    \subfigure[$V(\mathbf{x})$]{\label{fig:treed_gp_branin_var}
        \includegraphics[width=0.323\linewidth]{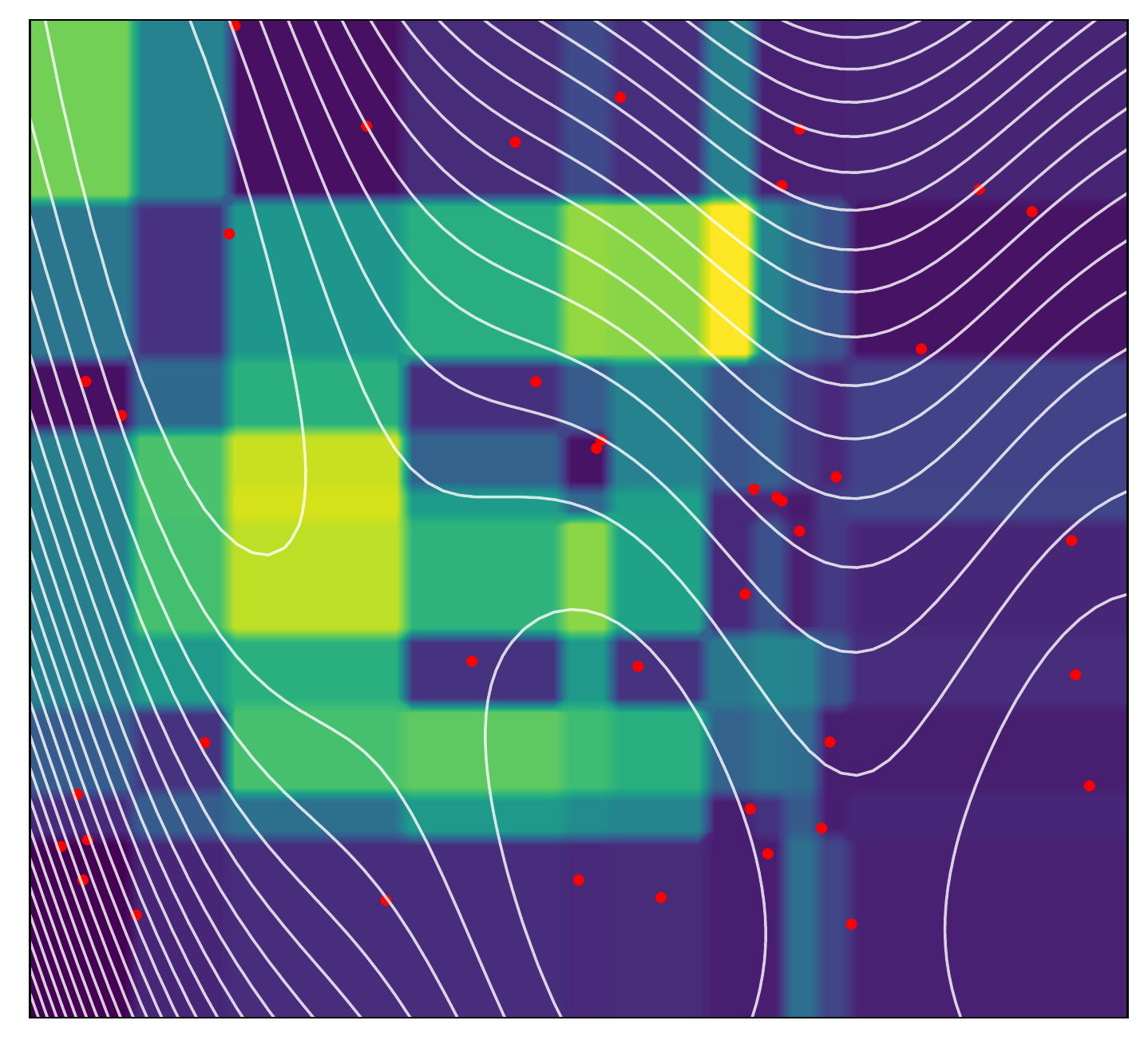}}
    \subfigure[$UCB(\mathbf{x})$]{\label{fig:treed_gp_branin_acq}
        \includegraphics[width=0.323\linewidth]{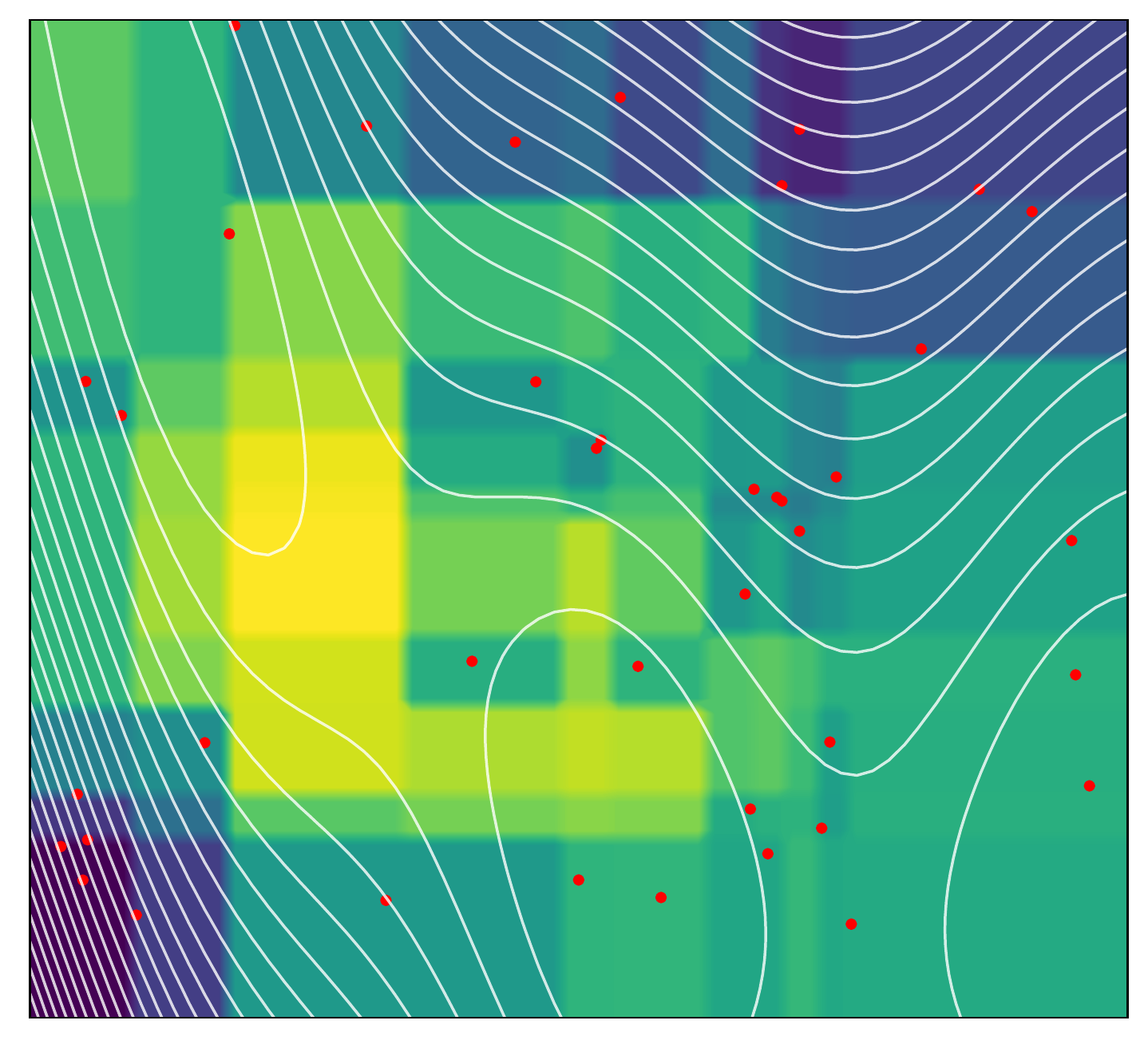}}
    \caption{Tree kernel GP trained on 40 random points of negative one times the Branin function for the intervals
        $\mathbf{x} \in (\left[ -5.0, 10.0 \right], \left[ 0.0, 15.0 \right])^{\intercal}$. 
        Function values increase with the colour brightness and white contour lines indicate the 
            true functional shape of the negated Branin function.}
    \label{fig:treed_gp_branin}
\end{figure}
Fig.~\ref{fig:treed_gp_branin} visualizes the Eq.~\eqref{eq:gp_infer} $M(\mathbf{x})$, 
    $V(\mathbf{x})$ and the upper-confidence bound (UCB) \citep{cox1992statistical} response surface.
Fig.~\ref{fig:treed_gp_branin_mean} shows that the tree kernel-based GP mean gives a good piecewise-constant approximation of negative one times the Branin function.
Fig.~\ref{fig:treed_gp_branin_var} shows that variance peaks reveal areas where data are
    sparse, reliably identifying uncertainty in the 
    underlying surrogate model.
Finally, Fig.~\ref{fig:treed_gp_branin_acq} shows how an acquisition function such as the UCB can effectively manage the exploitation-exploration trade-off. 

\subsection{Global optimization of tree ensembles}
\citet{misic2017OptimizationEnsembles} proposes a mixed-integer linear optimization formulation that ensures that binary variables $z_{t,l}$ follow the logic of the tree ensemble:
\begin{subequations}
    \label{eq:tree_constr}
    \begin{flalign}
        \sum\limits_{l\in{\mathcal{L}_{t}}} z_{t,l} &= 1, 
            &\forall t\in \mathcal{T}, 
            &\label{eq:tree_a}\\
        \sum\limits_{l\in\mathbf{left}(s)}\; z_{t,l} &\leq
            \sum\limits_{j \in \mathbf{C}(s)} \nu_{\text{V}(s),j},
            &\forall t \in \mathcal{T}, \forall s \in \mathbf{splits}(t),
            &\label{eq:tree_b}\\
        \sum\limits_{l\in\mathbf{right}(s)} z_{t,l} &\leq
            1 - \sum\limits_{j \in \mathbf{C}(s)} \nu_{\text{V}(s),j},
            &\forall t \in \mathcal{T}, \forall s \in \mathbf{splits}(t), 
            &\label{eq:tree_c} \\
        \sum\limits_{j=1}^{K_i} \nu_{i,j} &= 1,
            &\forall i \in \mathcal{C}, \label{eq:tree_d} \\
        \nu_{i,j} &\leq \nu_{i,j+1}, 
            &\forall i \in \mathcal{N}, \forall j \in \left [ K_i - 1 \right ],
            &\label{eq:tree_e} \\
        \nu_{i,j} &\in \{ 0,1  \},
            &\forall i \in \left [ n \right ], \forall j \in \left [ K_i \right ], 
            &\label{eq:tree_f} \\
        z_{t,l} &\geq 0, 
            &\forall t\in{\mathcal{T}}, \forall l\in{\mathcal{L}_{t}}. 
            &\label{eq:tree_g}
    \end{flalign}
\end{subequations}
Eq.~\eqref{eq:tree_a} guarantees exactly one active leaf $l$ in leaf set $\mathcal{L}_t$ for tree $t$.
Eqs.~\eqref{eq:tree_b}--\eqref{eq:tree_c} ensure that binary variables $z_{t,l}$ are only active if all previous split binaries $\nu_{\text{V}(s),j}$ corresponding to continuous splitting thresholds are active.
At any node in a given tree, $\mathbf{left}(s)$ and $\mathbf{right}(s)$ contain all leaves following the left and right branches, respectively.
The mapping $\text{V}(s)$ gives the splitting feature at node $s \in \mathbf{splits}(t)$ in tree $t$, with $\mathbf{splits}(t)$ defining the set of all splits in tree $t$.
Tree ensembles can handle continuous, integer and categorical data.
Continuous splits $s$ are defined by $x_{\text{V}(s)} \leq v_{\text{V}(s),j}$ conditions, where $v_{\text{V}(s),j}$ is the learned splitting threshold.
Therefore, $\mathbf{C}(s)$ only contains a single index $j$ representing the threshold $v_{\text{V}(s),j}$.
Categorical splits $s$ are characterized by (subsets of) categories available for feature $\text{V}(s)$ and define a splitting condition based on the inclusion of $x_{\text{V}(s)}$ in the category subset at split $s$. 
For categorical splits, $\mathbf{C}(s)$ includes the category indices  comprising the category subset at split $s$.
Eq.~\eqref{eq:tree_d} ensures that only one category is active per categorical variable $i \in \mathcal{C}$.
Continuous splitting thresholds of all trees in the ensemble are ordered according to $v_{i,1} < v_{i,2} < ... < v_{i,K_i}$ with $K_i$ denoting the index for the last split of continuous feature $i \in \mathcal{N}$.
To enforce this order, Eq.~\eqref{eq:tree_e} ensures that binary variables $\nu_{i,j}$, corresponding to the split thresholds $v_{i,j}$, are activated sequentially.
The model comprising Eqs.~\eqref{eq:tree_constr} has no direct dependency on $\mathbf{x}$ and is fully defined by binary variables indicating which splits and leaves of the tree model active.
However, to allow the user to include extra equality and inequality constraints on the input vector $\mathbf{x}$, we bound the continuous variables based on the active splits by adding linking constraints \citep{mistry2018MixedIntegerEmbedded}:
\begin{subequations}
    \label{eq:linking_constr}
    \begin{flalign} 
        x_{i} &\geq v^L_{i} + \sum\limits_{j=1}^{K_{i}} \left (v_{i,j} - v_{i,j-1} \right ) 
            \left ( 1 - \nu_{i,j} \right ), & &\forall i \in \mathcal{N},\\
        x_{i} &\leq v^U_{i} + \sum\limits_{j=1}^{K_{i}} \left (v_{i,j} - v_{i,j+1} \right ) 
            \nu_{i,j}, & &\forall i \in \mathcal{N},\\
        x_{i} &\in \left [ v_{i}^{L},v_{i}^{U} \right ], & &\forall i \in \mathcal{N},\\
        x_{i} &= \{j \in \left[K_i\right] \; | \; \nu_{i,j} = 1\}, 
            & &\forall i \in \mathcal{C},\label{eq:cat_map}
    \end{flalign}
\end{subequations}
For categorical variables, Eq.~\eqref{eq:cat_map} maps the indices of active categories onto $x_i$ for $i \in \mathcal{C}$. These optimization formulations are implemented in open-source software ENTMOOT \cite{thebelt2021entmoot} and OMLT \cite{ceccon2022omlt}.

\section{Tree ensemble kernels for Bayesian optimization} \label{sec:constrained_bayesian_optimization}

The technical details in Section \ref{sec:tree_kernel_inference} are insufficient to use the tree ensemble kernel in a Bayesian optimization framework. 
While Eqs.\ \ref{eq:tree_constr} and \ref{eq:linking_constr} allow optimization over the mean of an associated acquisition function \cite{misic2017OptimizationEnsembles}, Bayesian optimization also requires quantifying model uncertainty for exploration.
%
This section proposes a mixed-integer second-order cone optimization formulation to capture the standard deviation of a GP with a tree kernel prior. Combining this optimization formulation with the already-developed mixed-integer formulation of the mean function, we derive the upper-confidence bound (UCB) of the tree kernel-based GP. The advantage of deriving a mixed-integer second-order cone optimization formulation of the UCB acquisition function is that we can globally optimize the acquisition function. Additionally, we can easily incorporate explicit input constraints that capture domain knowledge and/or known search space relationships. 
    
Our optimization problem, which includes the UCB acquisition function, is:
\allowdisplaybreaks{
\begin{subequations}
    \begin{align}
    \label{eq:acq_func}
    \mathbf{x}_\text{lb}^*, \mathbf{x}_\text{ub}^*, \mathbf{x}_\text{cat}^* &\in 
    \underset{\mathbf{x}} {\text{arg max}} \;
    \mu(\mathbf{x}) + \kappa \sigma(\mathbf{x}), \\
    h(\mathbf{x}) &= 0, \\
    g(\mathbf{x}) &\leq 0,     
    \end{align}
\end{subequations}}
with $\mu(\mathbf{x})$ and $\sigma(\mathbf{x})$ denoting the surrogate model's mean prediction and 
    standard deviation, respectively.
Functions $h(\mathbf{x})$ and $g(\mathbf{x})$ are known constraints and handled similarly to \cite{Boukouvala2014,Tran2019}. In our implementation, the constraints can be linear, quadratic, or polynomial.
Hyperparameter $\kappa \geq 0$ controls the exploitation-exploration trade-off to determine the next black-box function query area. 
The solution to Eq.\ \ref{eq:acq_func} is defined by $\mathbf{x}_\text{lb}^*$ and $\mathbf{x}_\text{ub}^*$ for non-categorical variables $i \in \mathcal{N}$, i.e.,\ continuous and integer features, and $\mathbf{x}_\text{cat}^*$, a set of valid category subsets for categorical variables $i \in \mathcal{C}$. 
Two vectors ($\mathbf{x}_\text{lb}^*$ and $\mathbf{x}_\text{ub}^*$) define the non-categorical variables because the trees are piecewise-constant over intervals and the vectors define the lower and upper bounds of these intervals.
    
Next, we formalize the mean and variance of the tree kernel-based GP. Eqs.\ \ref{eq:kernel_constr} and \ref{eq:gp_infer} are equivalent.
\begin{subequations}
    \label{eq:kernel_constr}
    \begin{align}
        \mu(\mathbf{x}) = M(\mathbf{x}) &= K_{\mathbf{x},\mathbf{X}} \; (K_{\mathbf{X},
            \mathbf{X}})^{-1} \; \mathbf{y} & \label{eq:kernel_a} \\
        \sigma^2(\mathbf{x}) = 
        V(\mathbf{x}) &= K_{\mathbf{x},\mathbf{x}} - K_{\mathbf{x},\mathbf{X}} \;
            (K_{\mathbf{X},\mathbf{X}})^{-1} \; K_{\mathbf{x},
            \mathbf{X}}^{\intercal} & 
            &\label{eq:kernel_c}
    \end{align}
\end{subequations}
The Gram matrix $K_{\mathbf{X},\mathbf{X}}$ and target vector $\mathbf{y}$ are constants in the optimization model, as these quantities only depend on the data set $\mathbf{X}$. 
The value $K_{\mathbf{x},\mathbf{x}}$ is directly related to signal variance hyperparameter $\sigma_0$ since there is full leaf overlap for two identical inputs:
\begin{equation}
    K_{\mathbf{x},\mathbf{x}} = \sigma_0^2 \label{eq:kernel_d} 
\end{equation}
The vector $K_{\mathbf{x},\mathbf{X}}$ contains the kernel output of $\mathbf{x}$ with individual data points $\mathbf{x}_i$.
We compute the constant matrix $A \in \mathbb{R}^{m \times |\mathcal{L}|}$ after the gradient-boosted tree is trained but before solving the acquisition function. The entries of $A$ are equal to $1$ for all active leaves $l$ of data point $i$ and $0$ otherwise.
Given matrix $A$, Eqs.~\eqref{eq:kernel_constr_2} capture the kernel output by summing over binary variables $z_{t,l}$ that are active for data point $i$.
\begin{subequations}
    \label{eq:kernel_constr_2}
    \begin{align}
        K_{\mathbf{x},\mathbf{X}} &= \left[ k_\text{Tree}(\mathbf{x}, \mathbf{x}_1), 
            k_\text{Tree}(\mathbf{x}, \mathbf{x}_2), \dotsc, k_\text{Tree}(\mathbf{x},
            \mathbf{x}_m)\right] & 
            &\label{eq:kernel_e} \\
        k_\text{Tree}(\mathbf{x}, \mathbf{x}_i) &= \sigma_0^2 \; |\mathcal{T}|^{-1}
            \sum\limits_{t\in{\mathcal{T}}}\sum\limits_{l\in \mathcal{L}_t} A_{i,l} z_{t,l}
            & \forall i \in \left[m\right] 
            &\label{eq:kernel_f}
    \end{align}
\end{subequations}
To get an intuition for Eqs.\ \eqref{eq:kernel_constr_2}, note that the largest possible value for each $k_\text{Tree}(\mathbf{x}, \mathbf{x}_i)$ is $\sigma_0^2$ (if the corresponding entries of matrix $A$ are all equal to 1, and there is full leaf overlap) and the smallest possible value for each $k_\text{Tree}(\mathbf{x}, \mathbf{x}_i)$ is 0 (if there is no leaf overlap). In general, values of each element of vector $K_{\mathbf{x},\mathbf{X}}$ will range between 0 and $\sigma_0^2$: higher values in the elements of $K_{\mathbf{x},\mathbf{X}}$ indicate a higher degree of overlap between the next query location $\mathbf{x}$ and data point $\mathbf{x}_i$ in the set $\mathbf{X}$.

Without considering additional tree model constraints $h$ and $g$, the resulting acquisition function is a mixed-integer quadratic optimization problem. The optimization problem is \emph{mixed-integer} because of binary variables $\boldsymbol{\nu}$ and \emph{quadratic}
because of Eq.~\eqref{eq:kernel_c}. The quadratic Eq.~\eqref{eq:kernel_c} components are $\sigma^2$ and the terms $k_\text{Tree}^2(\mathbf{x}, \mathbf{x}_i)$ arising from the inner product of $K_{\mathbf{x},\mathbf{X}}$ with itself. 
We only require one direction of the Eq.~\eqref{eq:kernel_c} equality ($\leq$) and re-write Eq.~\eqref{eq:kernel_c} as a second-order cone constraint. Second-order cone programming \citep{alizadeh2003second, kuo2004interior, lobo1998applications} optimizes over a linear objective subject to both linear and second-order cone constraints (here, convex quadratic constraints, but the theory is more general).
More recently, solvers integrate advanced methods solving second-order cone problems in the mixed-integer setting \citep{drewes2009mixed, benson2013mixed, lubin2016extended}.
Solver Gurobi~9 \citep{gurobi9} automatically finds that Eq.~\eqref{eq:kernel_c} (with $\leq$ rather than $=$) can be represented as a second-order cone and makes the appropriate algorithm modifications, e.g.,\ as described by \cite{hijazi2014outer,vielma2017extended}. The proposed formulation is also compatible with open-source solver alternatives including Bonmin \citep{BONAMI2008186}, MindtPy \citep{bernal2018mixed}, Pajarito \cite{coey2020outer}, and SHOT \citep{lundell2022polyhedral}.
    
The acquisition function, with Objective \ref{eq:acq_func} and Constraints
 \eqref{eq:tree_constr}--\eqref{eq:kernel_f}, is a mixed-integer second-order cone program which can be solved with optimization solvers.
A valid solution to the proposed model is a set of active leaves $\mathcal{L}^*$ and the intersection of all leaf subspaces $\left[ \mathbf{x}_\text{lb}^*, \mathbf{x}_\text{ub}^* \right], \mathbf{x}_\text{cat}^*$.

To derive the next black-box function query point $\mathbf{x}^*$, we propose some heuristics.
The acquisition function value is constant for $\boldsymbol{x} \in \left[ \mathbf{x}_\text{lb}^*, \mathbf{x}_\text{ub}^* \right], \mathbf{x}_\text{cat}^*$, i.e.,\ all contained points are equivalent from the perspective of the tree kernel-GP.
For the continuous and integer features, we observe that tree models tend to learn split thresholds close to training data points. We propose the center of $\left[ \mathbf{x}_\text{lb}^*, \mathbf{x}_\text{ub}^* \right]$ as the next query point $\mathbf{x}^*$ for continuous and integer features: 
\begin{equation}
    \label{eq:leaf_center_unconstr}
        x_{\text{mid},i}^* = \frac{1}{2} \left( x_{\text{lb},i}^* + x_{\text{ub},i}^* \right), \forall i \in \mathcal{N}.
\end{equation}
For integer features with a fractional mid-point between the upper and lower bound, we randomly select its floor or ceiling. For categorical features, we sample from the subset of available categories:
\begin{equation}
        x_{\text{mid},i}^* = \mathrm{uniform}(\mathbf{x}_\text{cat}^*), \forall i \in \mathcal{C}.
        \label{eq:cat_query}
\end{equation}
For the unconstrained case, we use $\mathbf{x}_\text{mid}^*$ as the next query point $\mathbf{x}^*$.
When additional input constraints $h(\mathbf{x})$ and/or $g(\mathbf{x})$ are given, we know that at least one point in $\left[ \mathbf{x}_\text{lb}^*, \mathbf{x}_\text{ub}^* \right], \mathbf{x}_\text{cat}^*$ is feasible even if the heuristic $\mathbf{x}_\text{mid}^*$ is infeasible.
To repair the solution $\mathbf{x}_\text{mid}^*$, we project it onto the feasible space:
\begin{subequations}
    \label{eq:feas_center}
    \begin{flalign}
        \mathbf{x}^* \in &\underset{\mathbf{x} \in \left[ \mathbf{x}_\text{lb}^*, \mathbf{x}_\text{ub}^* \right], \mathbf{x}_\text{cat}^*}{\text{arg min}} \;
            \sum\limits_{i \in \mathcal{N}} \left( x_{\text{mid},i}^* - x_i \right)^2 - 
            \sum\limits_{i \in \mathcal{C}} \sum\limits_{j \in x_{\text{mid},i}^*} \nu_{i,j} & &\\
        &h(\mathbf{x}) = 0, & &\\ 
        &g(\mathbf{x}) \leq 0. & &
    \end{flalign}
\end{subequations}
Eq.~\eqref{eq:feas_center} projects $\mathbf{x}_\text{mid}^*$ onto the feasible set defined by $h(\mathbf{x})$ and $g(\mathbf{x})$.
The time complexity of solving Eq.~\ref{eq:feas_center}, and that of solving Eqs.~\eqref{eq:tree_constr}--\eqref{eq:kernel_constr} for the search space $\left[ \mathbf{x}_\text{lb}^*, \mathbf{x}_\text{ub}^* \right], \mathbf{x}_\text{cat}^*$, are both NP-hard. 
However, in preliminary evaluations we found that Gurobi~9 often solves both problems to $\epsilon$-global optimality for moderately-sized tree models in less time compared to random sampling. 

\textbf{Hyperparameters.}
We train the kernel hyperparameters
signal variance $\sigma_0^2$ and noise $\sigma_y^2$ 
by maximizing the log marginal likelihood. 
Additional hyperparameters are introduced by the gradient-boosted tree ensemble trained at every iteration, i.e.,\ maximum tree depth and number of trees, and through $\kappa$ in the UCB.
For the Section~\ref{sec:numerical_studies} numerical studies, we leave $\kappa$ and gradient-boosting hyperparameters constant and only increase maximum tree depth and number of trees for the high-dimensional CIFAR-NAS benchmark to capture more complicated interactions.
The appendix reports specific values for all hyperparameters.

\textbf{Limitations.}
The method suffers from standard BO challenges where the tree kernel may not be a good prior for the underlying black-box function, e.g.,\ purely continuous feature spaces.
More limitations arise from solving an NP-hard problem to global optimality when working with large data sets in high-dimensional search spaces.
For cases where solving the NP-hard problem is too difficult but BO is still applicable, Gurobi~9 can typically develop good feasible solutions as a heuristic.

\section{Numerical studies} \label{sec:numerical_studies}
This section empirically evaluates the performance of tree kernel-based GPs using a wide variety of synthetic and real-world benchmark problems.
We show (i) the ability of tree kernels to capture uncertainty of the underlying tree ensemble, (ii) the advantage of using global vs. local strategies for optimizing the acquisition function and (iii) the proposed algorithm's superior performance in cases with constrained search spaces and mixed variable types.
\texttt{LEAF-GP} denotes the Section~\ref{sec:constrained_bayesian_optimization} proposed algorithm, and Section~\ref{sec:related_work} outlines the baseline of methods we compare against.
For every run, we visualize the median and confidence intervals of the first and third quartile based on 20 individual runs with varying random seeds.
Further technical details can be found in the appendix. 

\subsection{Uncertainty metric} \label{sec:uncertainty_metric}
\begin{figure}[ht]
    \centering
    \subfigure[Relative Model Error]{\label{fig:unc_rastrigin_err}
        \includegraphics[height=3.8cm]{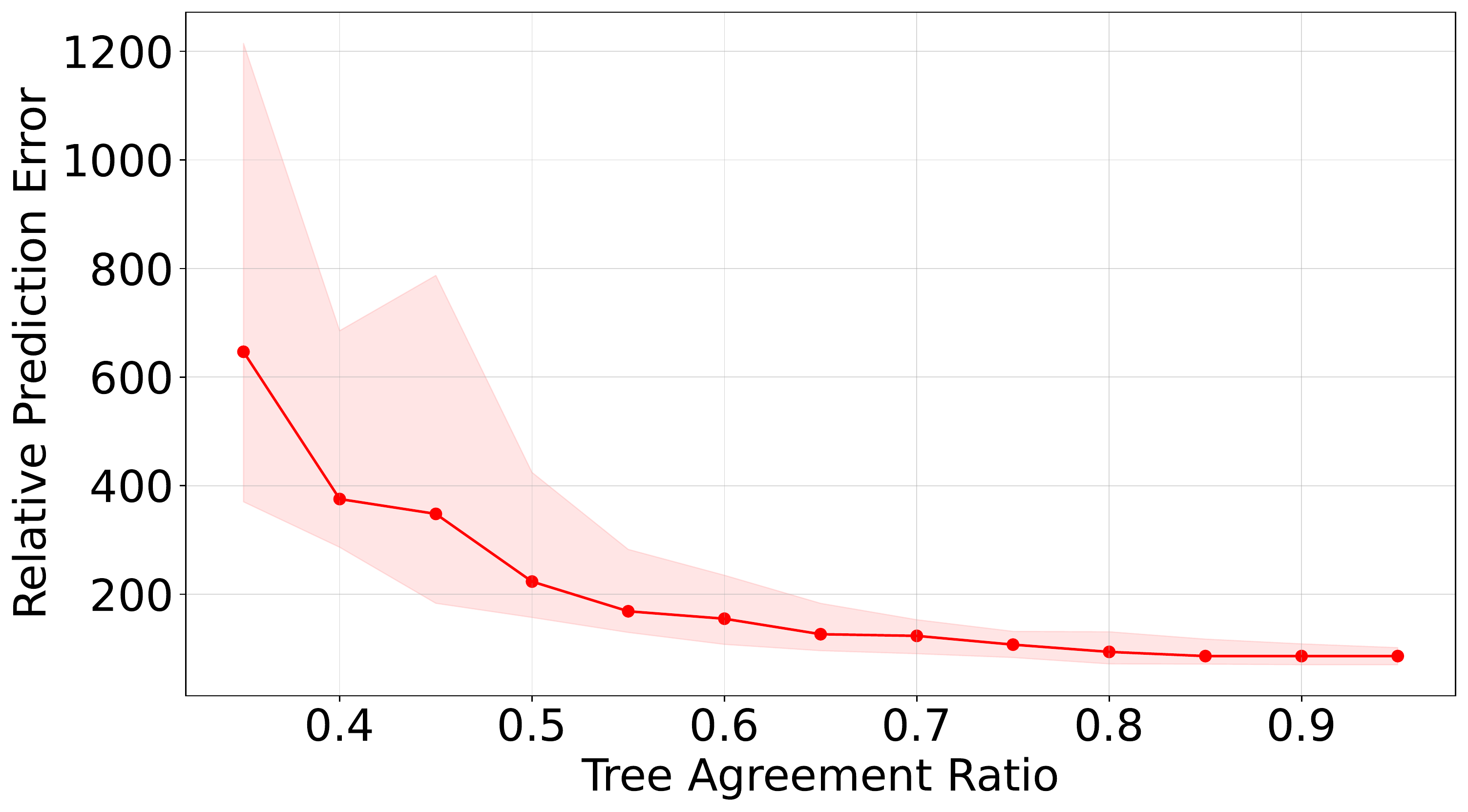}}
    \subfigure[Prediction Mean]{\label{fig:unc_rastrigin_mean}
        \includegraphics[height=3.8cm]{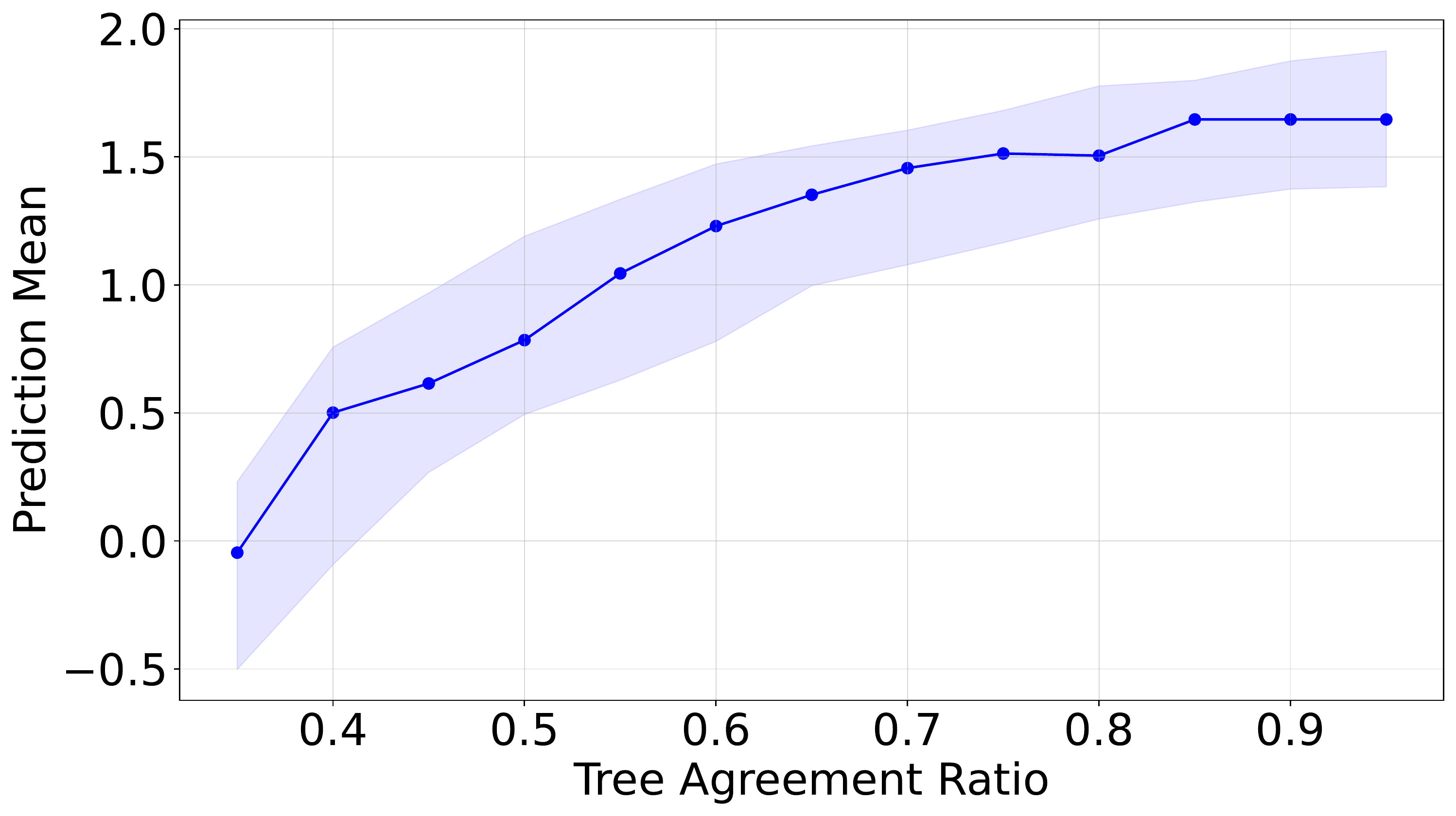}}
    \caption{The relative prediction error (Eq.~\ref{eq:model_error}) and model prediction mean over the maximum tree agreement ratio $R$ for benchmark problem Rastrigin (10D). Changing $R$ is equivalent to changing the maximum kernel covariance. Plot shows the median line and confidence intervals (first and third quartile) from 20 random seeds. Section~\ref{sec:uncertainty_metric} provides more details.}
    \label{fig:unc_rastrigin}
\end{figure}
The tree kernel-based GP uses the leaf overlap measure to quantify correlation between two inputs.
In BO, such measures help identify unexplored areas where correlation to existing training data is low and we expect inaccurate model predictions.
To empirically test the tree kernel's capability of identifying uncertainty in the underlying tree ensemble, we change the optimization formulation:
\begin{subequations}
    \label{eq:unc_study}
    \begin{flalign}
        \mathbf{x}_\text{lb}^*, \mathbf{x}_\text{ub}^*, \mathbf{x}_\text{cat}^* \in 
            &\underset{\mathbf{x}, \mathbf{z}, \mathbf{\nu}} {\text{arg max}} \;
            \mu(\mathbf{x}), \label{eq:unc_study_a} \\
        \text{s.t.} \; &\text{Eq.~\eqref{eq:tree_constr},
        Eq.~\eqref{eq:linking_constr},
        Eq.~\eqref{eq:kernel_constr},
        Eq.~\eqref{eq:kernel_d}, Eq.~\eqref{eq:kernel_constr_2}} \\
        &{|\mathcal{T}|}^{-1} \sum\limits_{t\in{\mathcal{T}}}\sum\limits_{l\in \mathcal{L}_t} A_{i,l} z_{t,l}
            \leq R, \; \forall i \in \left[m\right] \label{eq:unc_study_c}
    \end{flalign}
\end{subequations}
Eq.~\eqref{eq:unc_study_a} maximizes over the mean prediction of the tree kernel-based GP with Eq.~\eqref{eq:unc_study_c} restricting the ratio of tree agreement for the black-box query point $\mathbf{x}^*$ defined as the Eq.~\eqref{eq:leaf_center_unconstr} center of the optimal area $\mathbf{x}_\text{lb}^*, \mathbf{x}_\text{ub}^*, \mathbf{x}_\text{cat}^*$.
A solution to problem Eq.~\eqref{eq:unc_study} guarantees a maximum leaf overlap of $R$ between $\mathbf{x}^*$ and all available data points $\mathbf{X}$.
Limiting the tree agreement ratio corresponds to constraining the Eq.~\eqref{eq:tree_kernel} kernel correlation between the existing dataset and the proposed optimal area $\mathbf{x}_\text{lb}^*, \mathbf{x}_\text{ub}^*, \mathbf{x}_\text{cat}^*$.
Here, we evaluate values of $R \in \left[ 0.35, 1.0\right]$ with increments of $0.05$ and compute the model error according to:
\begin{equation}
    \label{eq:model_error}
    \epsilon_\text{error} = \left| \frac{\mu(\mathbf{x}^*) - f_\text{true}(\mathbf{x}^*)}{\mu(\mathbf{x}^*)} \right|
\end{equation}
Fig.~\ref{fig:unc_rastrigin} shows results for the Rastrigin \citep{simulationlib} benchmark.
With smaller values of $R$, the Eq.~\eqref{eq:unc_study} optimization problem becomes more restricted, leading to smaller objective values for solution $\mathbf{x}^*$. 
Fig.~\ref{fig:unc_rastrigin_err} shows that a growing leaf overlap (increasing $R$) reduces the model error defined in Eq.~\eqref{eq:model_error}.
This suggests that the kernel works as intended and that low kernel correlation can reveal search space areas with high model uncertainty.
The appendix provides results for additional benchmarks.

\subsection{Local vs. global acquisition optimization} \label{sec:local_vs_global}
\begin{figure}[ht]
    \centering
    \subfigure[Hartmann (6D)]{\label{fig:hartmann6d}
        \includegraphics[height=4.43cm]{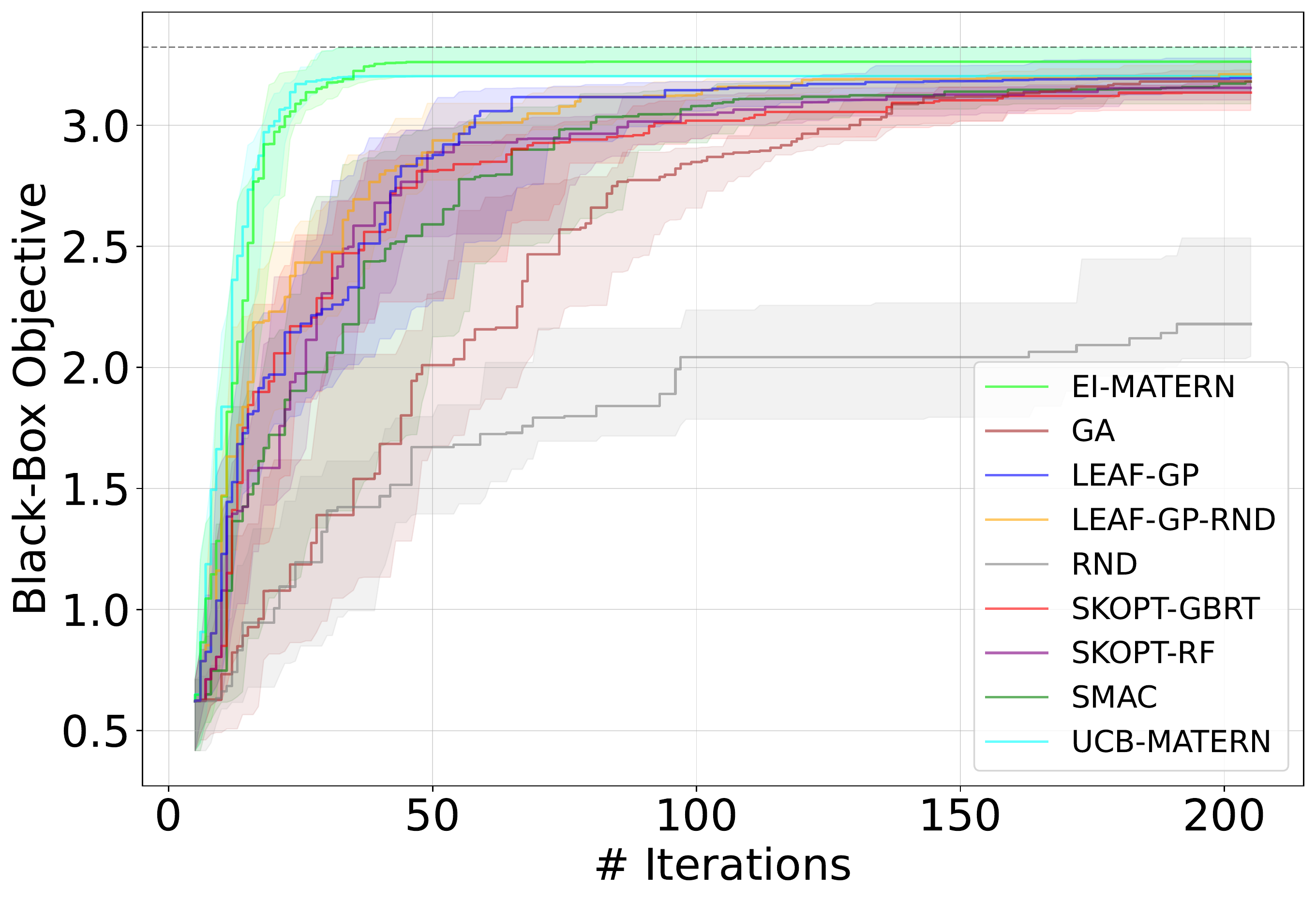}}
    \subfigure[Styblinski-Tang (10D)]{\label{fig:styblinski_tang}
        \includegraphics[height=4.43cm]{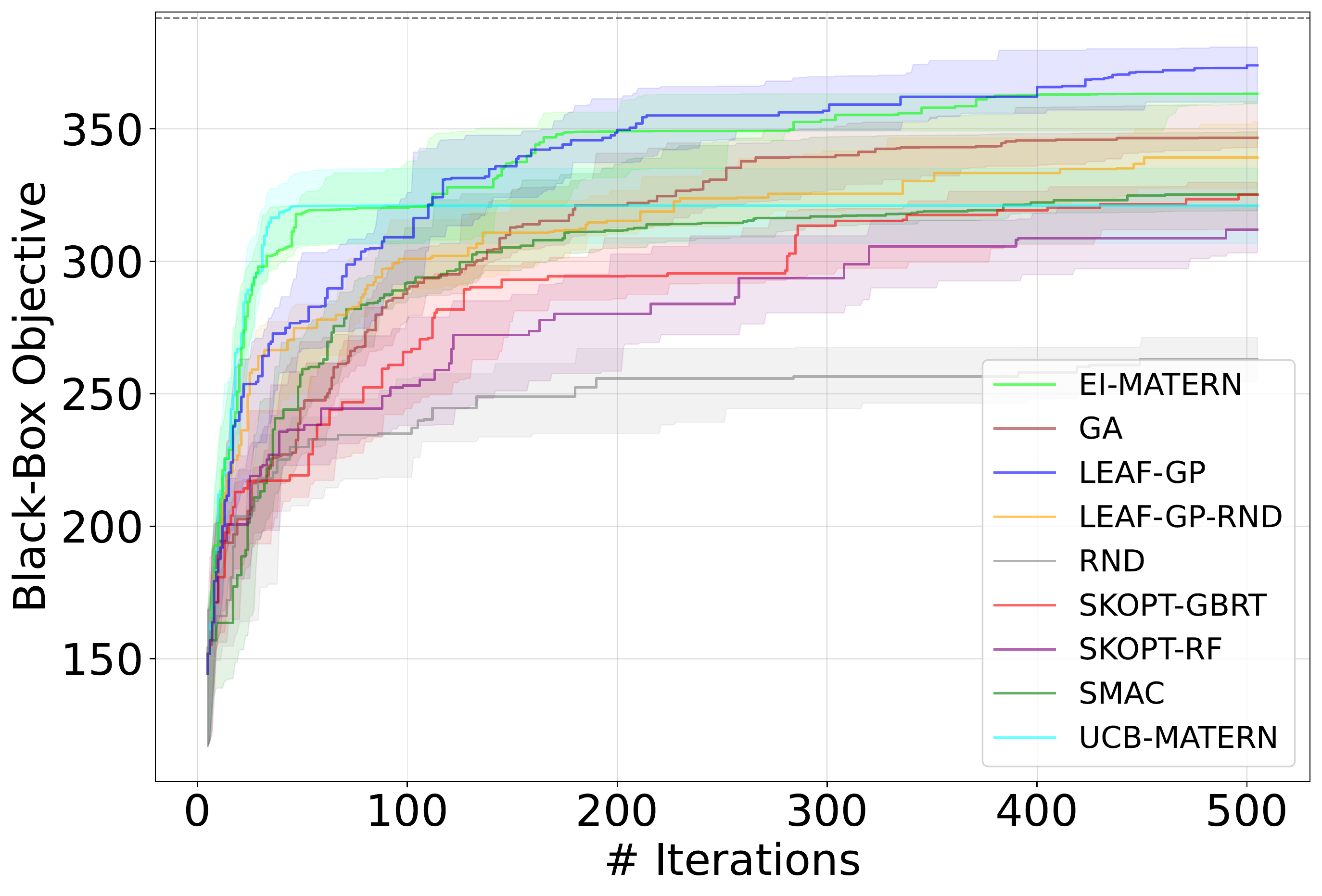}}   
    \caption{Black-box optimization progress of \texttt{LEAF-GP} vs.\ baseline. Plot shows the median line and confidence intervals (first and third quartile) from 20 random seeds. Section~\ref{sec:local_vs_global} provides more details}
    \label{fig:basic_test}
\end{figure}
This section (i) compares \texttt{LEAF-GP} to other state-of-the-art algorithms for common benchmarks and (ii) shows the advantage of global vs.\ local strategies for optimizing the acquisition function.
We introduce a variation of the proposed algorithm, \texttt{LEAF-GP-RND}, which optimizes the same acquisition function as \texttt{LEAF-GP}, but uses random sampling instead of Gurobi 9.
Fig.~\ref{fig:basic_test} shows results for Hartmann (6D) \citep{simulationlib} and Styblinski-Tang (10D) \citep{simulationlib}.
We do not expect tree model-based approaches to perform well since both benchmark functions are continuous.
However, the benchmarks show that all approaches perform similarly for Hartmann (6D) with \texttt{UCB-MATERN} and \texttt{EI-MATERN} improving the black-box objective at the fastest rate.
For Styblinski-Tang (10D), \texttt{LEAF-GP} significantly outperforms other algorithms. Although \texttt{LEAF-GP} is not specialized to this setting, observe that it performs roughly equivalently to the state of the art on these continuous, unconstrained optimization problems.

Although sampling-based \texttt{LEAF-GP-RND} and optimization-based \texttt{LEAF-GP} perform similarly on the smaller Hartmann (6D), \texttt{LEAF-GP} is particularly strong on the higher-dimensional Styblinski-Tang (10D) benchmark function, and two additional benchmarks in the appendix.
Moreover, \texttt{LEAF-GP-RND} does not support explicit input constraints and performs particularly bad for BO with known constraints.
For the following constrained benchmarks, we remove \texttt{LEAF-GP-RND} from the comparison.
The appendix has additional details and results. 

\subsection{Constrained spaces} \label{sec:constr_space}
\begin{figure}[ht]
    \centering
    \subfigure[G1 (13D, 9IC)]{\label{fig:g1}
        \includegraphics[height=4.43cm]{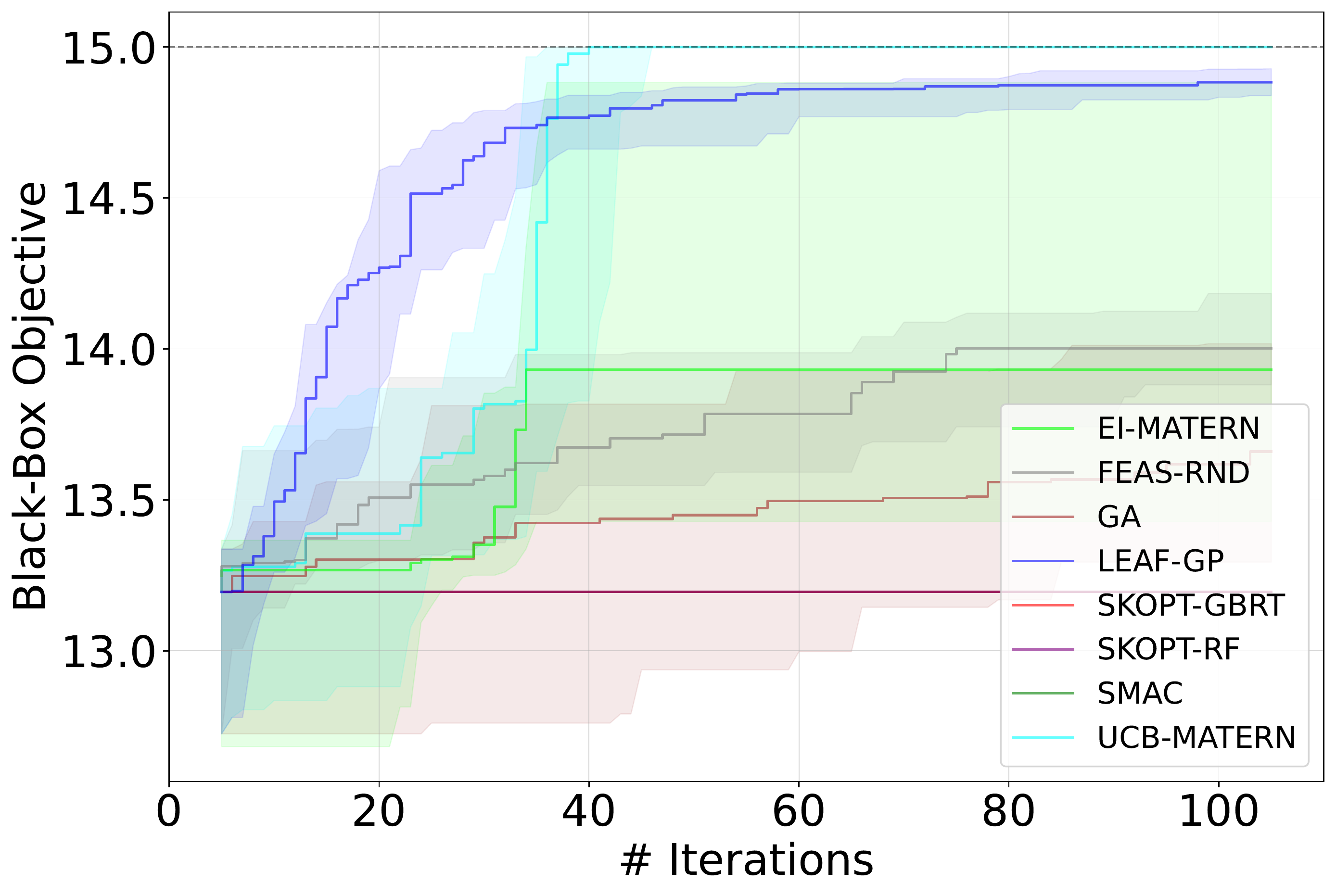}}
    \subfigure[G3 (5D, 1EC)]{\label{fig:g3}
        \includegraphics[height=4.43cm]{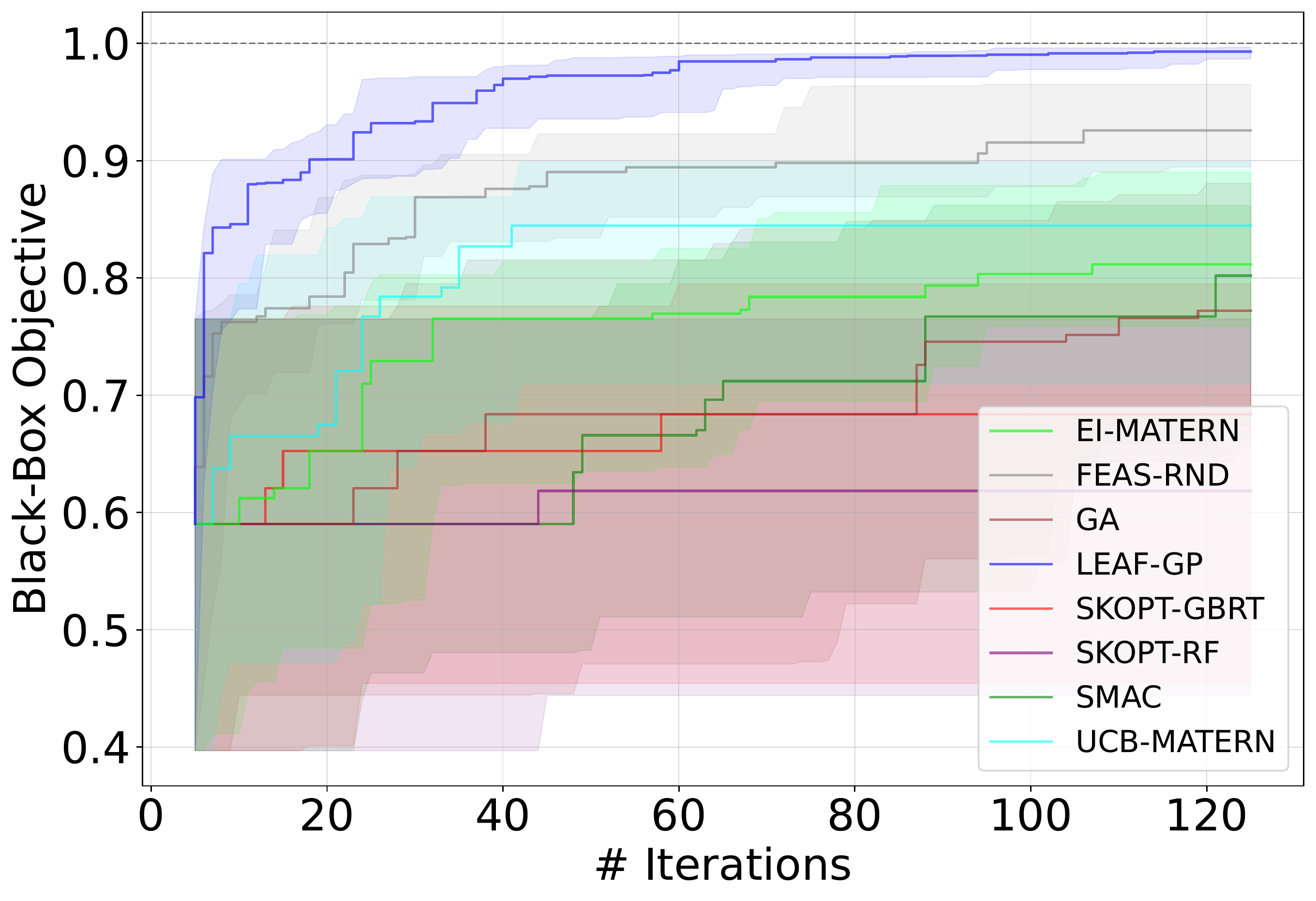}}
    \subfigure[G4 (5D, 6IC)]{\label{fig:g4}
        \includegraphics[height=4.43cm]{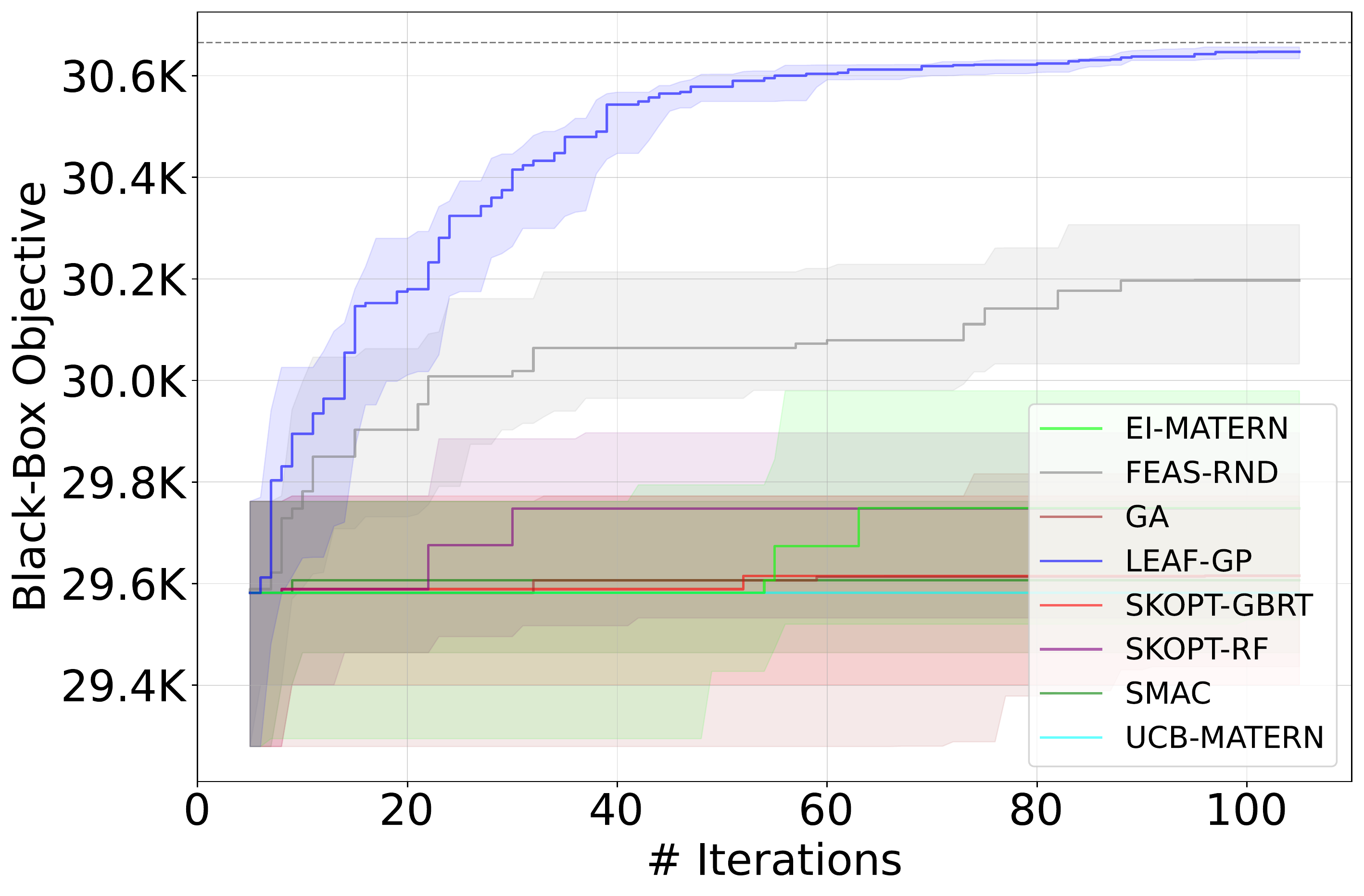}}
    \subfigure[Alkylation (7D, 14IC)]{\label{fig:alkyl}
        \includegraphics[height=4.43cm]{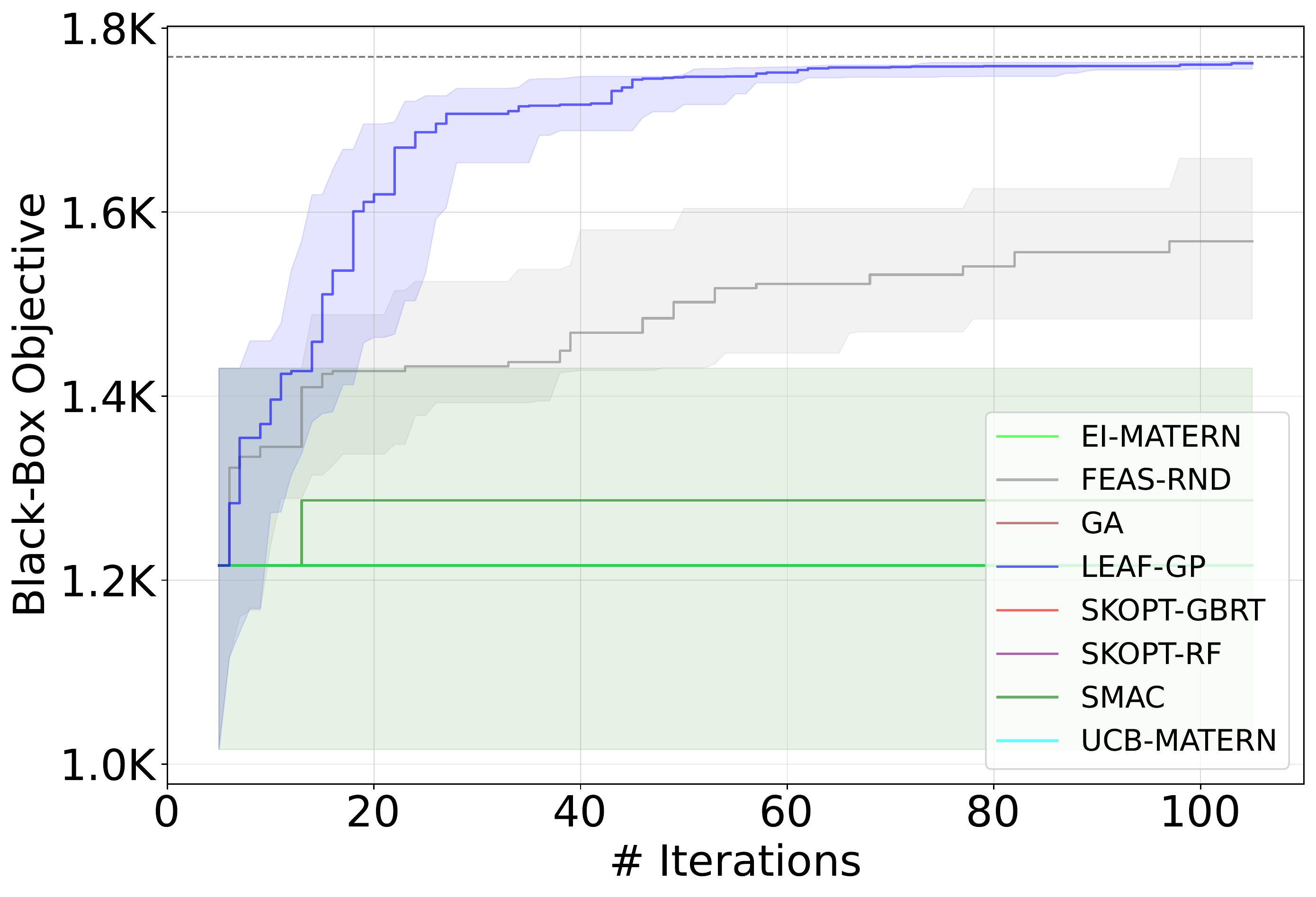}} 
    \caption{Feasible black-box optimization progress of \texttt{LEAF-GP} vs.\ baseline. Plot shows the median line and confidence intervals (first and third quartile) from 20 random seeds. Confidence intervals are neglected for methods that cannot improve the initial training data.
    Figure subtitles give the function name and number of: dimensions (D), equality constraints (EC), and inequality constraints (IC).
    Section~\ref{sec:constr_space} provides more details.}
    \label{fig:constr_test}
\end{figure}
This section presents numerical benchmarks with known input constraints.
The acquisition optimization strategy of \texttt{LEAF-GP} allows explicit consideration of input constraints, i.e.,\ logical and convex/non-convex $n$-th degree polynomial equality and inequality constraints for mixed variable spaces.
The \texttt{UCB-MATERN} and \texttt{EI-MATERN} implementations of BoTorch supports only linear equality and inequality constraints at the acquisition optimization step.
The \texttt{GA} pymoo algorithm has an interface for callable constraint functions which are considered when generating new generations of candidate points.
When a method does not support the specific input constraints, we penalize the objective:
\begin{equation}
    \label{eq:penalty}
    f_\text{penalty} = \lambda \left( \text{max}(g(\mathbf{x}), 0)^2 +  h(\mathbf{x})^2 \right),
\end{equation}
where $g(\mathbf{x}) \leq 0$ and $h(\mathbf{x}) = 0$ are inequality and equality constraints, respectively.
This penalty strategy allows methods that do not support explicit constraints to still produce feasible points given enough iterations.
Eq.~\eqref{eq:penalty} introduces the hyperparameter $\lambda$ which weights the penalty, we test values $\lambda \in \{1, 10, 100 \}$ and only plot the best run for all methods that rely on constraint penalization.
To initialize every method with feasible points, we draw random samples from a uniform distribution within the bounds that define the search space and compute the closest feasible point similar to Eq.~\eqref{eq:feas_center}.
For constrained benchmarks, we introduce \texttt{FEAS-RANDOM} which simply projects random samples onto the set of feasible points that satisfy $h(\mathbf{x})$ and $g(\mathbf{x})$.
Fig.\ \ref{fig:constr_test} plots feasible solutions to four different continuous benchmark problems, G1, G3, G4 \citep{constr_simulationlib}, and Alkylation \citep{sauer1964computer}.
The G1 benchmark has linear inequality constraints only, which are supported by \texttt{UCB-MATERN}, \texttt{EI-MATERN} and \texttt{LEAF-GP}.
Fig.~\ref{fig:g1} shows \texttt{LEAF-GP} making quick progress at the beginning with \texttt{UCB-MATERN} catching up towards the end.
G3 and G4 have different combinations of nonlinear constraints, and \texttt{LEAF-GP} significantly outperforms competing methods and random feasible sampling.
The Alkylation benchmark determines optimal operating conditions for a simplified alkylation process considering complicated nonlinear constraints representing economic, physical and performance limits.
Again, \texttt{LEAF-GP} outperforms other methods, which often struggle to find feasible solutions.
More details regarding the presented benchmark problems are given in the appendix.

\subsection{Mixed-variable spaces} \label{sec:mixed_vars}

\begin{figure}[ht]
    \centering
    \subfigure[Pressure Vessel (4D, 3IC)]{\label{fig:pressure_vessel}
        \includegraphics[height=4.43cm]{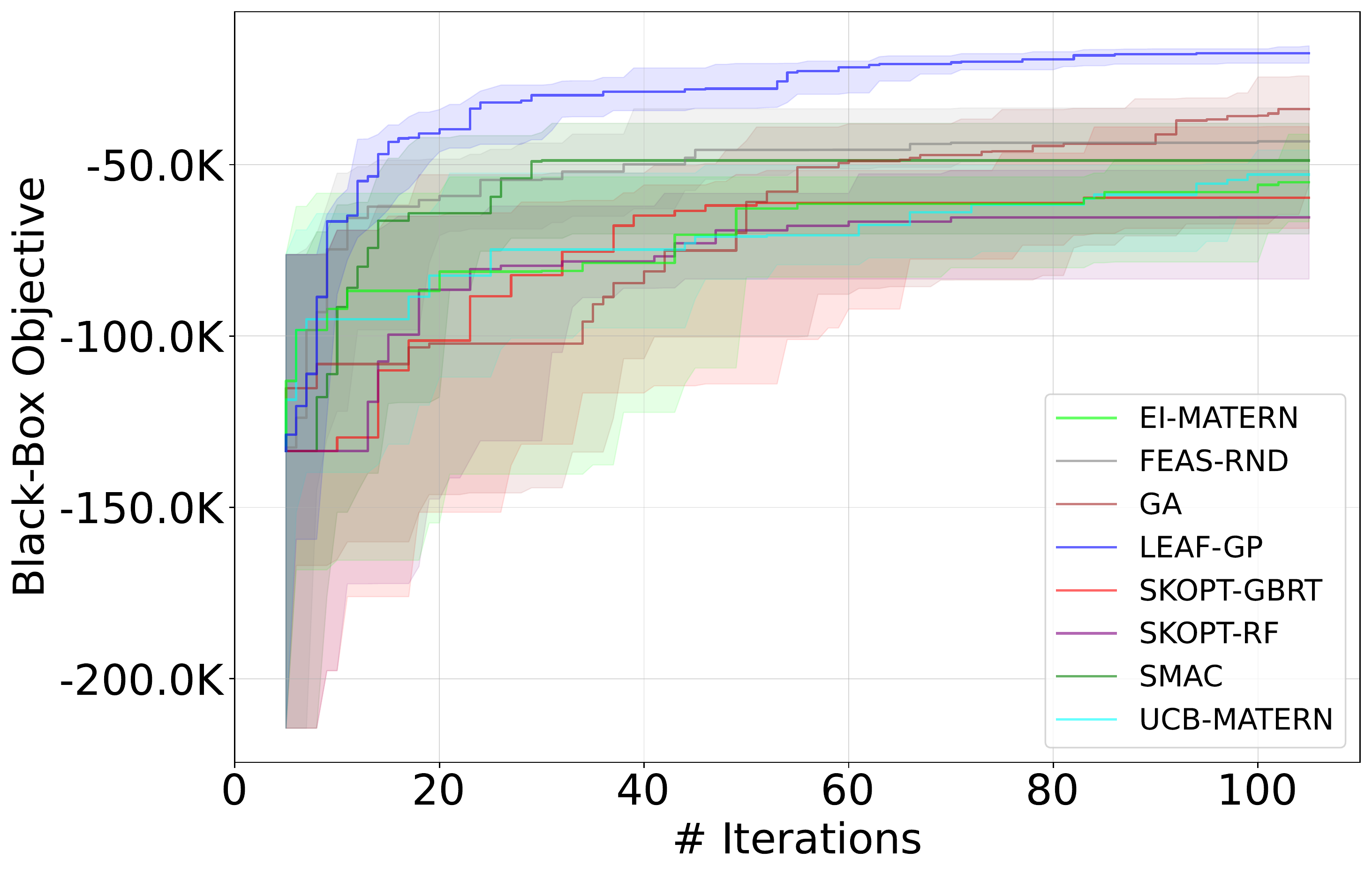}}   
    \subfigure[CIFAR-NAS (29D)]{\label{fig:cifar_small}
        \includegraphics[height=4.43cm]{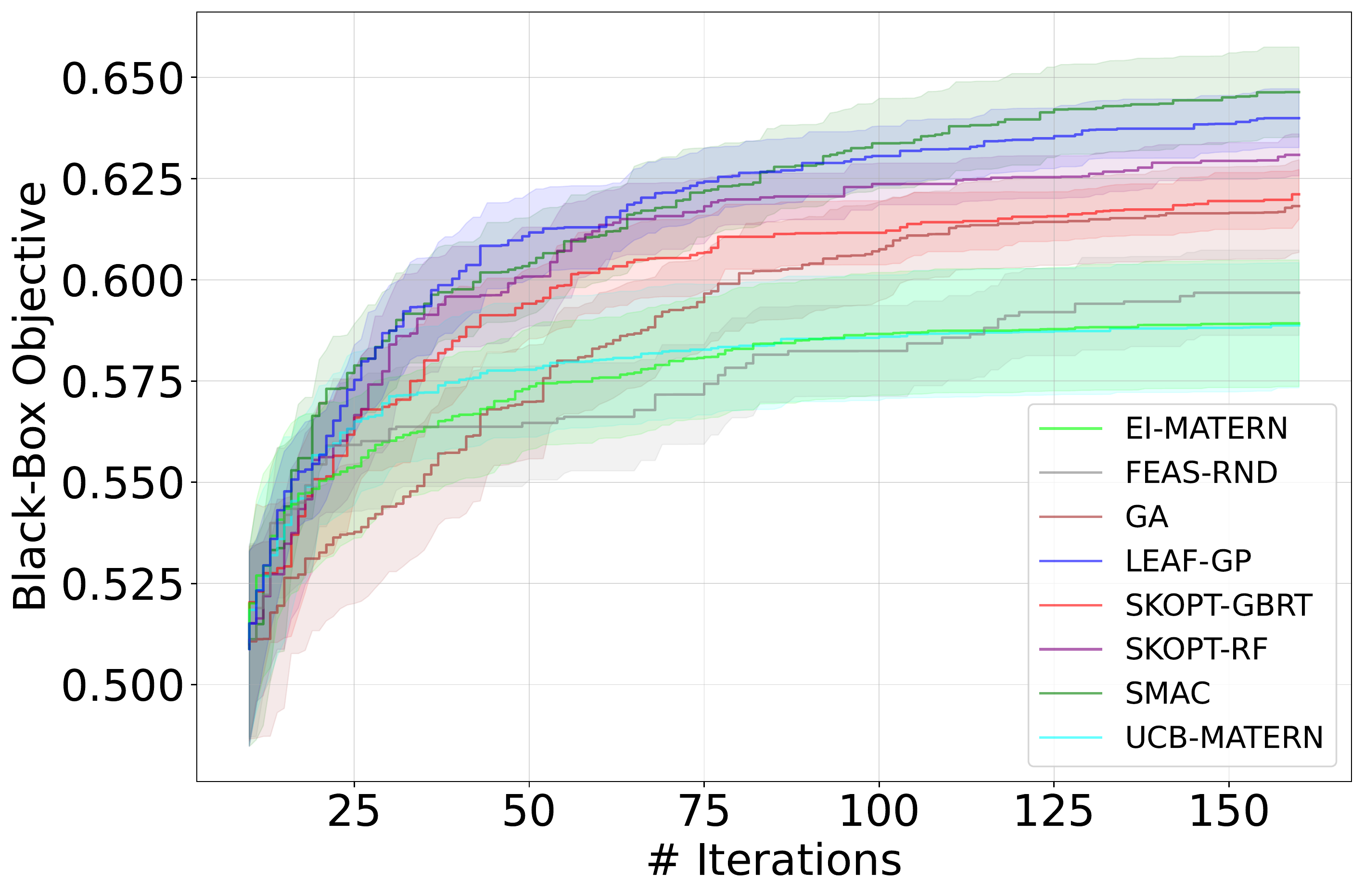}}  
    \caption{Feasible black-box optimization comparing \texttt{LEAF-GP} vs.\ baseline. Plot shows median line and confidence intervals (first and third quartile) from 20 random seeds. 
    Figure subtitles give the number of dimensions (D) and inequality constraints (IC).
    Section~\ref{sec:mixed_vars} provides details.}
    \label{fig:mixed_var_test}
\end{figure}

We now consider spaces that exhibit mixtures of continuous, integer and categorical variables for constrained problems.
Tree model-based algorithms naturally support categorical variables by replacing continuous splits with categorical splits.
For all methods that do not support categorical features, we use one-hot encoding.
The category with the highest corresponding auxiliary variable value is chosen for the subsequent query point.
On the Pressure Vessel (4D) \citep{coello2002constraint} benchmark, which comprises two continuous features, two integer features and three inequality constraints,
Fig.~\ref{fig:pressure_vessel} shows that \texttt{LEAF-GP} outperforms other black-box optimization methods.

The CIFAR-NAS (29D) is a high-dimensional benchmark problem with one continuous, 23 integer and five categorical variables. It describes properties of different layers and training hyperparameters for a convolutional neural network (CNN) trained on CIFAR-10 \citep{cifar10}.
The problem's search space is hierarchical, as different layers, i.e.,\ convolutional and fully-connected layers, can be activated.
Properties describing a layer are only relevant if the layer is active.
For this problem, we introduce two types of constraints to guide the \texttt{LEAF-GP}: (i) constraints that allow for feasible network designs, i.e.,\ different values for padding, stride, kernel size and max pooling in earlier layers affects what kernel size is feasible in following layers and (ii) constraints that ensure that hierarchies are respected.
Towards the former, we include constraints to ensure the output size of every convolutional layer is positive.
In general, tree models are particularly good at capturing hierarchical search space structures.
To enforce hierarchical relationships during the acquisition optimization step, we introduce indicator constraints that force layer properties to take predefined default values if the layer is inactive.
This ensures that the optimizer avoids leaves that infer splitting conditions from feature properties of inactive layers, thus effectively reducing the search space.
To expedite tests, we train the CNN on half of the CIFAR-10 training data and optimize for test accuracy.
Fig.~\ref{fig:cifar_small} summarizes the results of this study and shows that tree model-based algorithms generally outperform \texttt{UCB-MATERN} and \texttt{EI-MATERN}.
Utilizing search space constraints, \texttt{SMAC} and \texttt{LEAF-GP} find the neural architectures with the highest test accuracy with \texttt{SMAC} slightly outperforming \texttt{LEAF-GP}.
Since finding feasible architectures for CIFAR-10 is not particularly challenging, Fig.~\ref{fig:vae_small} in the appendix shows a benchmark for tuning variational autoencoders (VAE) adapted from \citet{daxberger2019mixed}, where \texttt{LEAF-GP} significantly outperforms other algorithms.
Finding feasible VAE architectures is more difficult given the requirement that the latent encoding must be decoded back to original input image size. 

\section{Conclusion}
We present a framework for black-box optimization based on tree kernel Gaussian processes that simultaneously allows (i) reliable uncertainty quantification of mixed feature spaces and (ii) incorporation of explicit input constraints. Although these two needs have been considered separately, we are able to address both simultaneously through the mixed-integer second-order cone formulation of the acquisition function.
The numerical studies show that the proposed strategy performs competitively with state-of-the-art algorithms for unconstrained problems and may significantly outperform existing methods for constrained benchmarks, especially those with mixed feature spaces.
We use optimization constraints together with the acquisition functions to incorporate  domain knowledge and leverage hierarchical search space structures, e.g.,\ for neural architecture search.

\section{Acknowledgements}
This work was supported by BASF SE, Ludwigshafen am Rhein, EPSRC Research Fellowships to R.M. (EP/P016871/1) and CT (EP/T001577/1), and an Imperial College Research Fellowship to CT.

\bibliography{citations}

\clearpage

\section*{Checklist}


\begin{enumerate}

\item For all authors...
\begin{enumerate}
  \item Do the main claims made in the abstract and introduction accurately reflect the paper's contributions and scope?
    \answerYes{See Abstract and Section~\ref{sec:motivation}.}
  \item Did you describe the limitations of your work?
    \answerYes{See Section~\ref{sec:constrained_bayesian_optimization}.}
  \item Did you discuss any potential negative societal impacts of your work?
    \answerNo{There is no obvious societal impact of the method proposed.}
  \item Have you read the ethics review guidelines and ensured that your paper conforms to them?
    \answerYes{}
\end{enumerate}

\item If you are including theoretical results...
\begin{enumerate}
  \item Did you state the full set of assumptions of all theoretical results?
    \answerNA{}
        \item Did you include complete proofs of all theoretical results?
    \answerNA{}
\end{enumerate}

\item If you ran experiments...
\begin{enumerate}
  \item Did you include the code, data, and instructions needed to reproduce the main experimental results (either in the supplemental material or as a URL)?
    \answerYes{We submit the code to run all experiments for the proposed method.}
  \item Did you specify all the training details (e.g., data splits, hyperparameters, how they were chosen)?
    \answerYes{See end of Section~\ref{sec:constrained_bayesian_optimization} and supplemental material.}
        \item Did you report error bars (e.g., with respect to the random seed after running experiments multiple times)?
    \answerYes{We run 20 random seeds for every experiment and report median and confidence intervals, i.e.,\ first and third quartile. See Section~\ref{sec:numerical_studies}.}
        \item Did you include the total amount of compute and the type of resources used (e.g., type of GPUs, internal cluster, or cloud provider)?
    \answerYes{We report the computational resources used to conduct the experiments in the supplementary material.}
\end{enumerate}

\item If you are using existing assets (e.g., code, data, models) or curating/releasing new assets...
\begin{enumerate}
  \item If your work uses existing assets, did you cite the creators?
    \answerYes{Yes, we cite all libraries that were used when preparing the numerical studies. For more information, we refer to the supplemental material.}
  \item Did you mention the license of the assets?
    \answerNo{License information for all assets can be found in the documentation of each of the assets.}
  \item Did you include any new assets either in the supplemental material or as a URL?
    \answerNo{The code to reproduce the paper's main results will be published open-source after the peer review process.}
  \item Did you discuss whether and how consent was obtained from people whose data you're using/curating?
    \answerNA{}
  \item Did you discuss whether the data you are using/curating contains personally identifiable information or offensive content?
    \answerNA{}
\end{enumerate}

\item If you used crowdsourcing or conducted research with human subjects...
\begin{enumerate}
  \item Did you include the full text of instructions given to participants and screenshots, if applicable?
    \answerNA{}
  \item Did you describe any potential participant risks, with links to Institutional Review Board (IRB) approvals, if applicable?
    \answerNA{}
  \item Did you include the estimated hourly wage paid to participants and the total amount spent on participant compensation?
    \answerNA{}
\end{enumerate}

\end{enumerate}

\clearpage

\appendix

\section{Appendix}

\section{General experimental setup}
All experimental results presented in Section~\ref{sec:numerical_studies} were evaluated on an HTCondor cluster (see \citep{thain2005Condor}) of machines equipped with Intel Core i7-8700 3.20GHz and 16 GB RAM.
Confidence intervals show the first and third quartile of 20 independent runs with random seeds $\in \left[101, 102, \dots, 120 \right]$.
If not specifically indicated in Section~\ref{sec:algorithms}, all competing algorithms use default values for all hyperparameters.
To allow for a fair comparison, we give the same set of initial data points to all tested methods for both constrained and unconstrained benchmark problems.
A set of ten initial samples is used for the Section~\ref{sec:mixed_vars} CIFAR-NAS example due to its high dimensionality.
Five initial points are used for the remaining benchmark problems. 

\section{Algorithms} \label{sec:algorithms}
This section summarizes the different algorithms used for the Section~\ref{sec:numerical_studies} numerical studies.
We give implementation details and hyperparameter settings to reproduce the presented results.

\subsection{LEAF-GP and LEAF-GP-RND}
The \texttt{LEAF-GP} uses LightGBM \citep{ke2017LightGBM} for training gradient-boosted tree ensembles.
All runs use the hyperparameter value $\textit{min\_data\_in\_leaf} = 1$ as the training dataset size needs to be at least twice the minimum number of data points a leaf is based on.
The $\textit{min\_data\_in\_leaf}$ default value of LightGBM is 20, which would cause run-time errors after initialization.
We also set LightGBM hyperparameter $\textit{min\_data\_per\_group} = 1$.
For the high-dimensional benchmark problem CIFAR-NAS, we set $\textit{max\_depth} = 5$ and $\textit{num\_boost\_rounds} = 100$ for training the ensemble in LightGBM, referring to maximum interaction depth per decision tree and total number of trees in the ensemble, respectively.
For all other benchmarks we use $\textit{max\_depth} = 3$ and $\textit{num\_boost\_rounds} = 50$.

We implement the tree ensemble kernel as a non-stationary kernel in GPyTorch \citep{gardner2018gpytorch}.
For deriving the posterior distribution, we use a Gaussian likelihood and standardize the target values of the data set.
Section~3 introduces signal variance $\sigma_0$ and noise term $\sigma_y$ as kernel hyperparameters which are fitted by maximizing the marginal log likelihood over 200 epochs using the Adam solver \citep{kingma2014adam}.
The hyperparameters are constrained by intervals according to $\sigma_0 \in \left[5\mathrm{e}{-4}, 0.2\right]$ and $\sigma_y \in \left[0.05, 20.0\right]$.

For \texttt{LEAF-GP} the Section~\ref{sec:constrained_bayesian_optimization} acquisition optimization formulation is encoded using \textit{gurobipy} \citep{gurobi9} and solved using Gurobi~9.
Runs are limited to 100 s if the solver finds a feasible solution and are continued otherwise.

Moreover, we set $\textit{heuristics} = 0.2$ and activate the non-convex hyperparameter for benchmark problems with non-convex constraints.
\texttt{LEAF-GP-RND} uses a sampling-based strategy that randomly evaluates the acquisition function at 2000 locations and selects the maximimum value.

\subsection{GA}
We use the standard Genetic Algorithm implementation of the pymoo \citep{pymoo} toolbox for evolutionary algorithms and change $\textit{population\_size} = 10$ given the small evaluation budget.
Default values are used for all other hyperparameters.

\subsection{SKOPT-GBRT and SKOPT-RF}
We use the default implementation of Scikit-Optimize \citep{head2018ScikitOptimize} with random forest and gradient-boosted trees base estimators for \texttt{SKOPT-RF} and \texttt{SKOPT-GBRT}, respectively.
Default values are used for all hyperparameters.

\subsection{SMAC}
For \texttt{SMAC} \citep{hutter2011SequentialModel} we utilize the most recent Python implementation SMAC3 \citep{lindauer2021smac3} using random forest models.
Moreover, we specify the hyperparameters $\textit{run\_obj} =$ `$\mathit{quality}$' and activate the deterministic flag to allow for reproducibility.
The Section~\ref{sec:mixed_vars} CIFAR-NAS and Section~\ref{app:vae_nas} VAE-NAS benchmark problems exhibit hierarchical search space relationships, i.e.,\ hyperparameters of a specific layer are only relevant if the layer is active.
We use \texttt{SMAC}'s \texttt{InCondition} function which allows for certain child features to be considered only if some parent features have certain values, e.g.,\ the stride of layer $n$ is only considered if number of layers is at least $n$.
This allows \texttt{SMAC} to capture hierarchical relationships explicitly and to avoid evaluating multiple equivalent configurations.
Default values are used for all other hyperparameters.

\subsection{UCB-MATERN and EI-MATERN}
The \texttt{UCB-MATERN} and \texttt{EI-MATERN} algorithms use the standard upper confidence bound and expected improvement implementations of BoTorch \citep{balandat2020botorch}.
Before fitting a GP, we normalize data features and standardize data outputs.
The upper confidence acquisition hyperparameter $\beta$ is set to 1.96.
We negate the target values and define $\textit{raw\_samples} = 200$ for the acquisition function maximization.
For unconstrained cases, the acquisition optimizer uses 100 restarts.
However, for constrained problems we limit the optimizer to five restarts due to extensive run-times caused by finding feasible solutions.
Default values are used for all other hyperparameters.

\section{Benchmark problems}

\subsection{Unconstrained problems}
Figure~\ref{fig:basic_test} shows the results for Hartmann (6D), Rastrigin (10D), Schwefel (10D) and Styblinski-Tang (10D) benchmark functions implemented according to \citet{simulationlib}.
Table~\ref{tab:basic_test} summarizes number of dimensions and domain evaluated for unconstrained benchmark problems.

\begin{table}[h]
  \caption{Benchmark functions for local vs. global acquisition-function optimization tests. Table shows function name, number of dimensions (Dim.), and domain of input variables}
  \label{tab:basic_test}
  \centering
  \begin{tabular}{lll}
    \midrule
    Function & Dim. & Domain \\
    \midrule
    Hartmann        & 6  & $\mathbf{x} \in \left[0.0, 1.0\right]^{6}$ \\
    Rastrigin       & 10 & $\mathbf{x} \in \left[-4.0, 5.0\right]^{10}$ \\
    Schwefel        & 10 & $\mathbf{x} \in \left[-500.0, 500.0\right]^{10}$ \\
    Styblinski-Tang & 10 & $\mathbf{x} \in \left[-5.0, 5.0\right]^{10}$ \\
    \bottomrule
  \end{tabular}
\end{table}

\subsection{Constrained problems}
\begin{figure}[ht]
    \centering
    \subfigure[G1 (13D, 9IC)]{\label{fig:app_g1}
        \includegraphics[height=4.43cm]{figures/g1.pdf}}
    \subfigure[G3 (5D, 1EC)]{\label{fig:app_g3}
        \includegraphics[height=4.43cm]{figures/g3.pdf}}
    \subfigure[G4 (10D, 6IC)]{\label{fig:app_g4}
        \includegraphics[height=4.43cm]{figures/g4.pdf}}
    \subfigure[G6 (2D, 2IC)]{\label{fig:app_g6}
        \includegraphics[height=4.43cm]{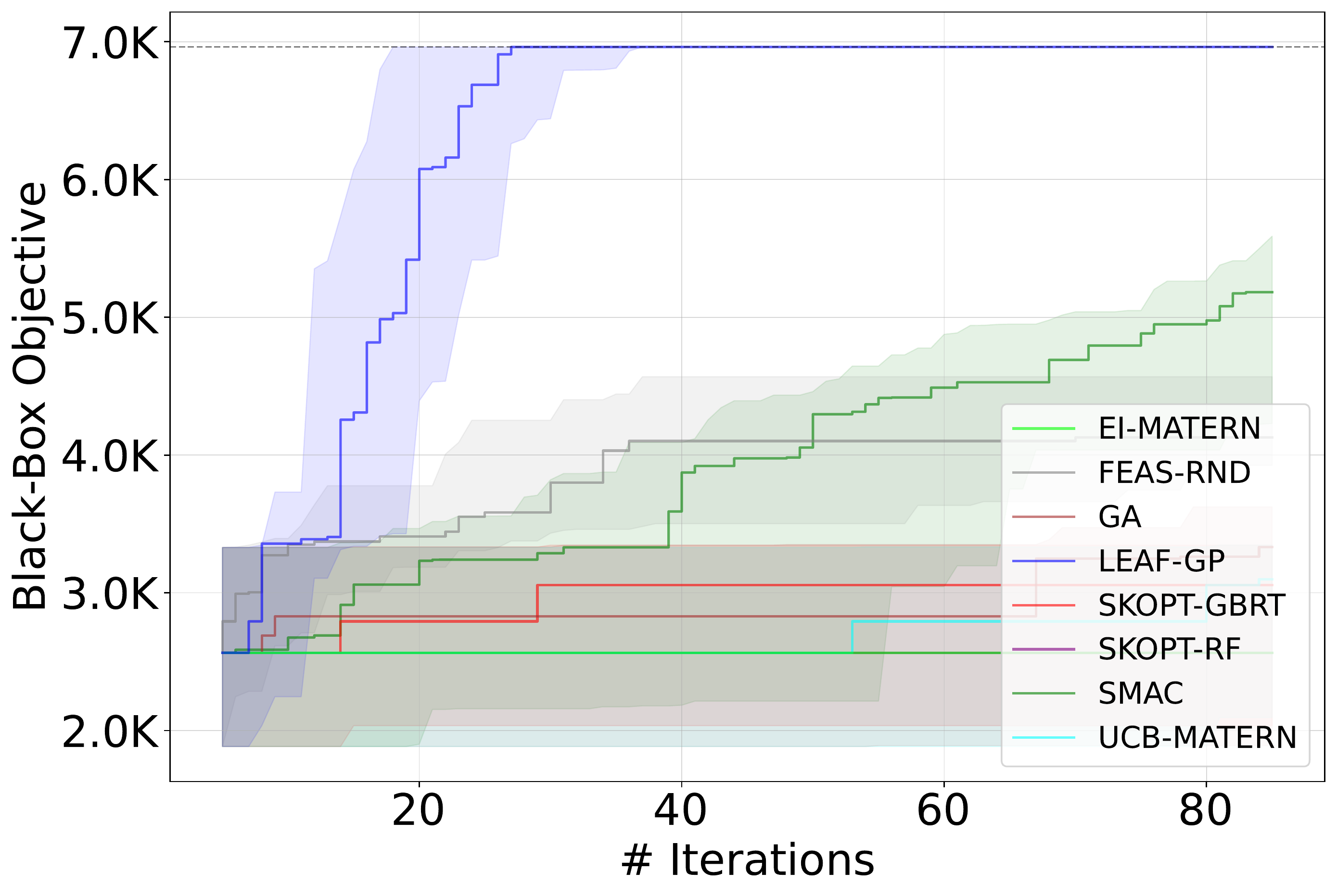}}   
    \subfigure[G7 (10D, 8IC)]{\label{fig:app_g7}
        \includegraphics[height=4.43cm]{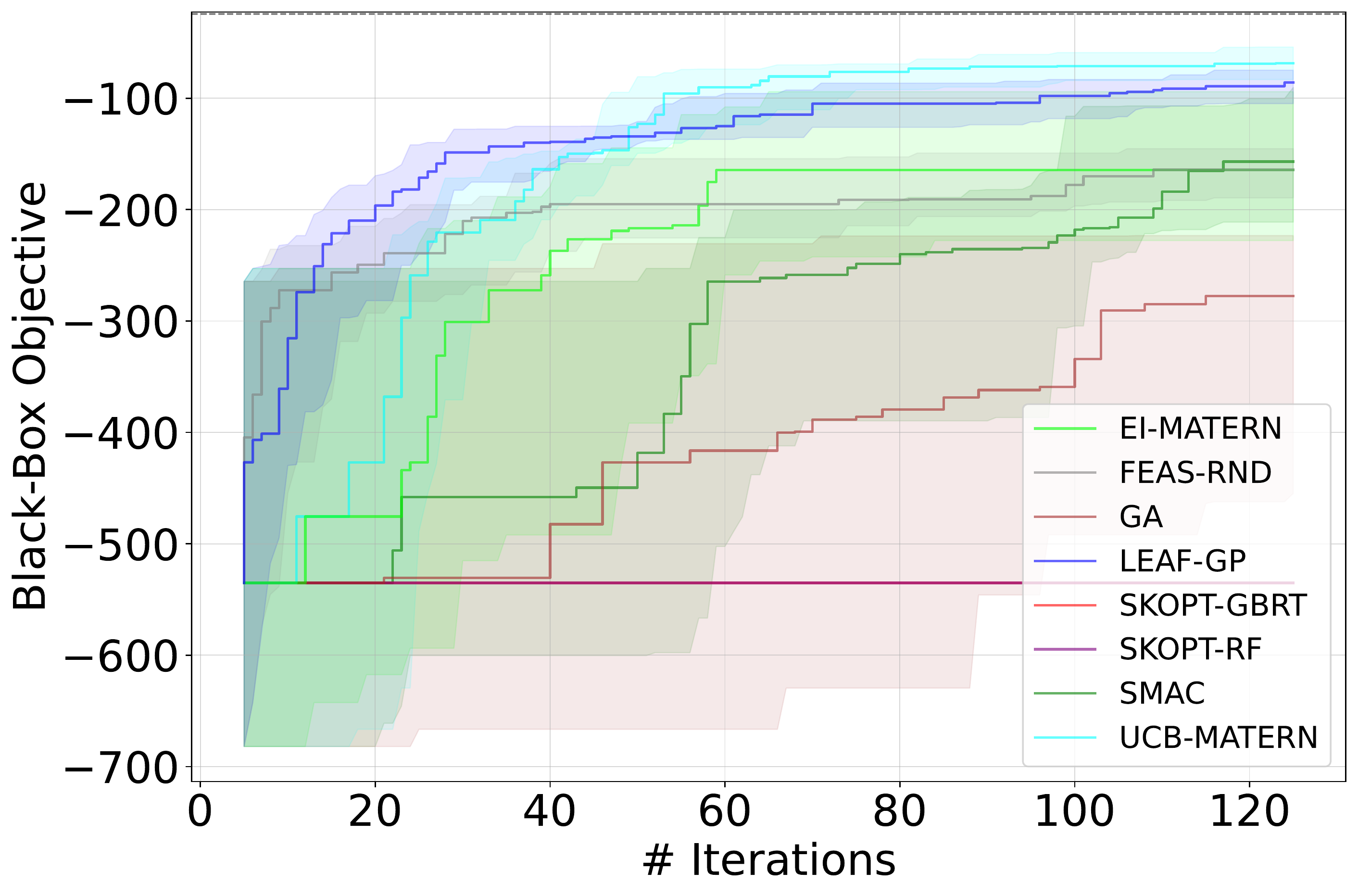}}   
    \subfigure[G10 (8D, 6IC)]{\label{fig:app_g10}
        \includegraphics[height=4.43cm]{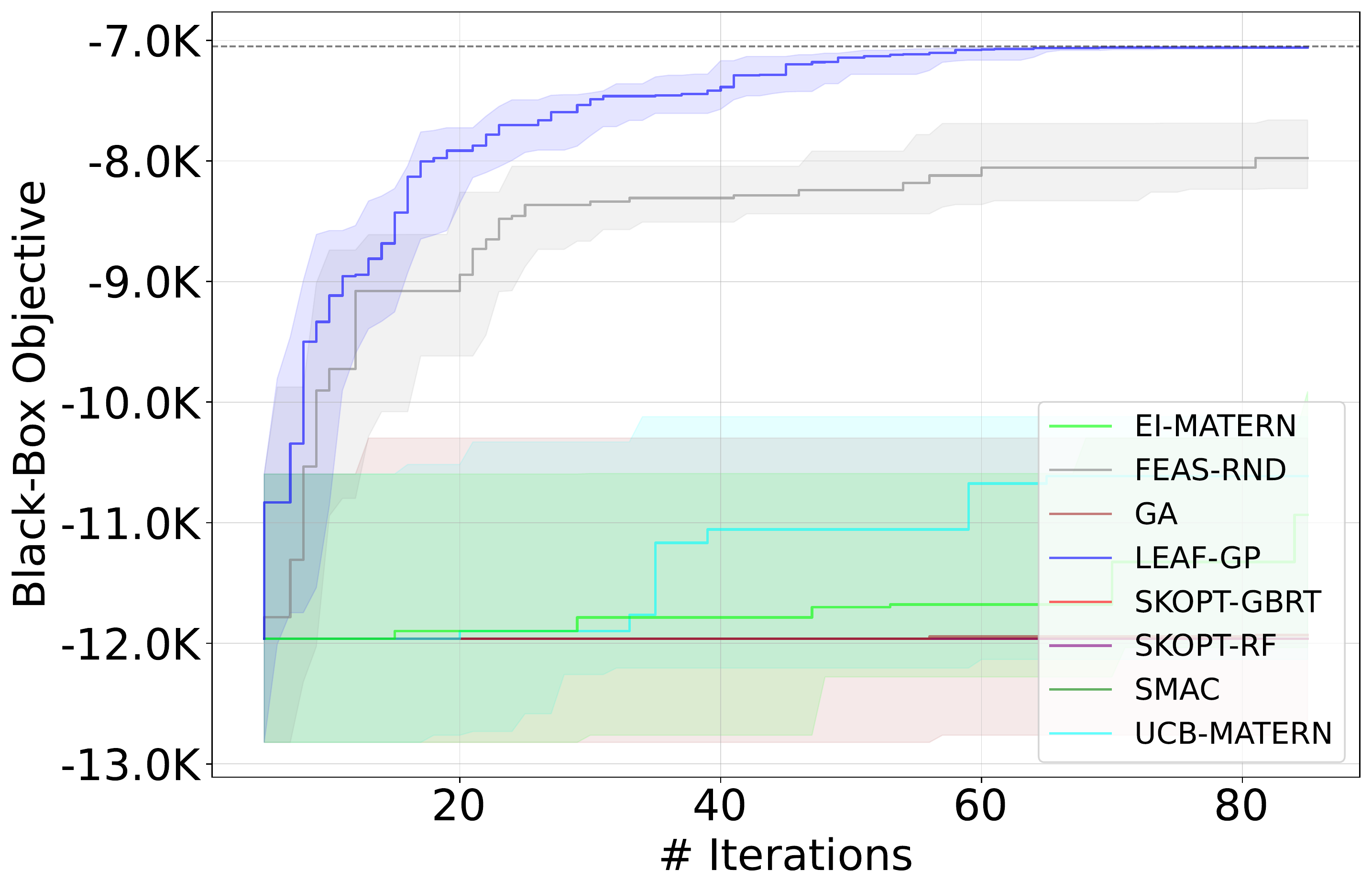}}   
    \subfigure[Alkylation (7D, 14IC)]{\label{fig:app_alkyl}
        \includegraphics[height=4.43cm]{figures/alkylation.pdf}}
    \subfigure[Pressure Vessel (4D, 3IC)]{\label{fig:app_pressure_vessel}
        \includegraphics[height=4.43cm]{figures/pressure_vessel.pdf}} 
    \caption{Feasible black-box optimization progress of \texttt{LEAF-GP} vs.\ baseline. Plot shows the median line and confidence intervals (first and third quartile) from 20 random seeds. Confidence intervals are neglected for methods that cannot improve the initial training data.
    Figure subtitles give the function name and number of: dimensions (D), equality constraints (EC), and inequality constraints (IC).
    Section~\ref{sec:constr_space} provides more details.}
    \label{fig:app_constr_test}
\end{figure}

Fig.~\ref{fig:app_constr_test} presents results of benchmark problems with known constraints.
Domain bounds without decimals indicate integer-valued variable types.
Benchmark examples G1, G3, G4, G6, G7 and G10 are implemented according to \citet{constr_simulationlib}.
The Alkylation benchmark \citep{sauer1964computer} is adapted from an open-source implementation \citep{alky_implementation}.
To compare methods that do not support specific input constraints, we penalize the black-box function output according to:
\begin{equation}
    \label{eq:app_penalty}
    f_\text{penalty} = \lambda \left( \text{max}(g(\mathbf{x}), 0)^2 +  h(\mathbf{x})^2 \right),
\end{equation}
where $g(\mathbf{x}) \leq 0$ and $h(\mathbf{x}) = 0$ are inequality and equality constraints, respectively.
Maximizing the combined black-box output allows methods to find feasible solutions.
Eq.~\eqref{eq:app_penalty} has the hyperparameter $\lambda$ which weights the penalty, we test values $\lambda \in \{1, 10, 100 \}$ and only plot the best run for all methods that rely on constraint penalization.
In the tests conducted, \texttt{LEAF-GP} fully supports explicit input constraints.
\texttt{UCB-MATERN} and \texttt{EI-MATERN} support linear inequality and equality constraints only.
The evolutionary algorithm \texttt{GA} has built-in constraint consideration but does not guarantee feasible solutions.
\texttt{LEAF-GP-RND}, \texttt{SKOPT-GBRT}, \texttt{SKOPT-RF} and \texttt{SMAC} rely on the Eq.~\eqref{eq:app_penalty} penalty function.

\begin{table}[h]
  \caption{Benchmark functions for constrained search space tests. Table shows function name and number of: dimensions (D), equality constraints (EC), and inequality constraints (IC). Values in brackets indicate the number of linear constraints which are natively supported by some of the algorithms. Domain bounds without decimals indicate integer-valued variables.}
  \label{tab:constr_test}
  \centering
  \begin{tabular}{lllll}
    \midrule
    Function & D & IC & EC & Domain \\
    \midrule
    G1       & 13 & 9 (9) & 0 (0) & \makecell[l]{
        $\mathbf{x}_{\{0, \dots, 8, 12\}} \in \left[0.0, 1.0\right]$, \\ 
        $\mathbf{x}_{\{9, 10, 11\}} \in \left[0.0, 100.0\right]$} \\
    \midrule
    G3       & 5 & 0 (0) & 1 (0)  & $\mathbf{x} \in \left[0.0, 1.0\right]^{5}$ \\
    \midrule
    G4       & 5 & 6 (0) & 0 (0)  & \makecell[l]{
        $\mathbf{x}_0 \in \left[78.0, 102.0\right]$,
        $\mathbf{x}_1 \in \left[33.0, 45.0\right]$, \\
        $\mathbf{x}_{\{2, 3, 4\}} \in \left[27.0, 45.0\right]$} \\
    \midrule
    G6       & 2 & 2 (0) & 0 (0)  & $\mathbf{x}_0 \in \left[13.0, 100.0\right]$,
        $\mathbf{x}_1 \in \left[0.0, 100.0\right]$ \\
    \midrule
    G7       & 10 & 8 (3) & 0 (0)  & $\mathbf{x} \in \left[-10.0, 10.0\right]^{10}$ \\
    \midrule
    G10        & 8 & 6 (3) & 0 (0)  & \makecell[l]{
        $\mathbf{x}_0 \in \left[100.0, 10.0\text{K}\right]$, \\
        $\mathbf{x}_{\{1, 2\}} \in \left[1.0\text{K}, 10.0\text{K}\right]$, \\
        $\mathbf{x}_{\{3, \dots 7\}} \in \left[10.0, 1.0\text{K}\right]$} \\
    \midrule
    Alkylation & 7 & 14 (0) & 0 (0)  & \makecell[l]{
        $\mathbf{x}_0 \in \left[0.0, 2.0\text{K}\right]$,
        $\mathbf{x}_1 \in \left[0.0, 16.0\text{K}\right]$, \\
        $\mathbf{x}_2 \in \left[0.0, 120.0\right]$, 
        $\mathbf{x}_3 \in \left[0.0, 5.0\text{K}\right]$, \\
        $\mathbf{x}_4 \in \left[90.0, 95.0\right]$,
        $\mathbf{x}_5 \in \left[0.01, 4.0\right]$, \\
        $\mathbf{x}_6 \in \left[145.0, 162.0\right]$}\\
    \midrule
    Pressure Vessel & 4 & 3 (2) & 0 (0)  & \makecell[l]{
        $\mathbf{x}_{\{0,1\}} \in \left[1, 99\right]$,
        $\mathbf{x}_{\{2,3\}} \in \left[10.0, 200.\right]$} \\
    \bottomrule
  \end{tabular}
\end{table}

\subsection{CIFAR-NAS} \label{app:cifar_nas}
Table~\ref{tab:cifar_test} gives more details on the Section~\ref{sec:mixed_vars} CIFAR-NAS (29D) benchmark problem.
CIFAR-NAS (29D) has a total of 29 hyperparameters to tune, i.e.,\ 1 continuous, 15 integer, 8 binary and 5 categorical variables.
The goal is to select hyperparameter values for a CNN trained in PyTorch \citep{NEURIPS2019_9015} on the CIFAR-10 dataset \citep{cifar10} that maximize test accuracy.
The training and test scripts were adapted from \citet{cifar_base}.
Due to limited computing resources, we train the CNN on half of the training data for 10 epochs using the Adam solver \citep{kingma2014adam}.
We score networks using the full test data set. 
Only certain combinations of stride, padding and filter size for various layers result in feasible neural architectures, e.g.,\ the filter size of one layer may be too large given the output of the previous layer.
In such cases the CNN training fails, and the black-box returns the largest black-box value found so far, helping algorithms learn to avoid infeasible neural architectures.
To simplify the training, layer inputs are parameterized based on outputs of the previous layer for convolutional and fully-connected layers, as well as the intermediate connecting layer.
The benchmark introduces categorical variables for activation function selection.
Methods that do not support categorical features use one-hot encoding.

\texttt{LEAF-GP} has access to constraints capturing feasible neural architectures mainly concerned with the convolutional layers.
Algorithms can choose to activate at most three convolutional and two fully-connected layers.
To capture constraints for feasible CNNs, we introduce $w_{\text{out},i}$ as the output of convolutional layer $i$ and $W_{\text{out},i}$ as the layer's modified output in case max-pooling is applied:
\begin{subequations}
    \begin{flalign}
        w_{\text{out},i} &\in \mathbb{N}_0, \forall i \in \{1,2,3\} \\ 
        W_{\text{out},i} &\in \mathbb{N}_0, \forall i \in \{1,2,3\}
    \end{flalign}    
\end{subequations}

Convolutional layers use PyTorch's \texttt{Conv2D} with inputs derived by the optimization algorithms.
PyTorch's \texttt{MaxPool2d} implements the max-pooling with the commonly-used $(2,2)$ kernel size.

\begin{subequations}
    \label{eq:layer_feas}
    \begin{flalign}
        W_{\text{in},1} &= 32 \label{eq:layer_feas_a} \\
        w_{\text{out},1} &= \frac{W_{\text{in},1} - F_1 + 2P_1}{S_1} + 1 \label{eq:layer_feas_b} \\
        W_{\text{out},1} &= b_1^\text{conv}\lfloor w_{\text{out},1} (1-0.5b^\text{pool}_1) \rfloor + (1-b_1^\text{conv})W_{\text{in},1} \label{eq:layer_feas_c} \\
        w_{\text{out},2} &= \frac{W_{\text{out},1} - F_2 + 2P_2}{S_2} + 1 \label{eq:layer_feas_d} \\
        W_{\text{out},2} &= b_2^\text{conv}\lfloor w_{\text{out},2} (1-0.5b^\text{pool}_2) \rfloor + (1-b_1^\text{conv})W_{\text{out},1} \label{eq:layer_feas_e} \\
        w_{\text{out},3} &= \frac{W_{\text{out},2} - F_3 + 2P_3}{S_3} + 1 \label{eq:layer_feas_f} \\
        W_{\text{out},3} &= b_3^\text{conv}\lfloor w_{\text{out},3} (1-0.5b^\text{pool}_3) \rfloor + (1-b_3^\text{conv})W_{\text{out},2} \label{eq:layer_feas_g} \\
        W_{\text{out},3} &\geq 1 \label{eq:layer_feas_h} \\
        1 &\leq b_1^\text{conv} + b_2^\text{conv} + b_3^\text{conv} + b_1^\text{fc} + b_2^\text{fc} \label{eq:layer_feas_i}
    \end{flalign}
\end{subequations}

The $32 \times 32$ image size of CIFAR-10 data defines the input to the full CNN ($W_{\text{in},1}$) in Eq.~\eqref{eq:layer_feas_a}.
Eq.~\eqref{eq:layer_feas_b} combines filter size $F_1$, padding $P_1$ and stride $S_1$ of the first convolutional layer to compute its output size $w_{\text{out},1}$.
Variable $W_{\text{out},1}$ captures the final output of the convolutional layer by considering if the layer is activated, i.e., $b_1^\text{conv} = 1$, and if max-pooling is applied, i.e.,\ $b^\text{pool}_1 = 1$.
Constraints~\eqref{eq:layer_feas_c}--\eqref{eq:layer_feas_g} denote the same restrictions for subsequent layers, each using the output size of the previous layer as its input size.
Eq.~\eqref{eq:layer_feas_h} ensures that the output of the last convolutional layer $W_{\text{out},3}$ is at least one.
We also enforce that at least one layer be active, which is captured by Eq.~\eqref{eq:layer_feas_i}.

To break symmetries in the benchmark problems, we introduce Constraints~\eqref{eq:layer_conv_cond_a}--\eqref{eq:layer_conv_cond_g}:

\begin{subequations}
    \label{eq:layer_conv_cond}
    \begin{flalign}
        b_3^\text{conv} &\leq b_2^\text{conv} \leq b_1^\text{conv} \label{eq:layer_conv_cond_a} \\
        \neg b_i^\text{conv} &\rightarrow \neg b_i^\text{pool}, \forall i \in \{1,2,3\} \label{eq:layer_conv_cond_b} \\
        \neg b_i^\text{conv} &\rightarrow C^\text{conv}_i \leq 4, \forall i \in \{1,2,3\} \label{eq:layer_conv_cond_c} \\ 
        \neg b_i^\text{conv} &\rightarrow F_i \leq 2, \forall i \in \{1,2,3\} \label{eq:layer_conv_cond_d} \\
        \neg b_i^\text{conv} &\rightarrow S_i \leq 1, \forall i \in \{1,2,3\} \label{eq:layer_conv_cond_e} \\
        \neg b_i^\text{conv} &\rightarrow P_i \leq 0, \forall i \in \{1,2,3\} \label{eq:layer_conv_cond_f} \\
        \neg b_i^\text{conv} &\rightarrow Act^\text{conv}_i = \text{ReLU}, \forall i \in \{1,2,3\} 
        \label{eq:layer_conv_cond_g}
    \end{flalign}
\end{subequations}

Eq.~\eqref{eq:layer_conv_cond_a} activates layers in a particular order, and Eq.~\eqref{eq:layer_conv_cond_b} deactivates max-pooling when the associated convolutional layer is inactive.
Constraints~\eqref{eq:layer_conv_cond_c}--\eqref{eq:layer_conv_cond_g} set layer-specific hyperparameters to pre-defined default values when the associated layer is inactive.
We select these defaults as the lower bound for non-categorical variables and the first category for categorical variables.

Constraints~\eqref{eq:layer_fully_cond} express the same restrictions for fully-connected layers:

\begin{subequations}
    \label{eq:layer_fully_cond}
    \begin{flalign}
        b_2^\text{fc} &\leq b_1^\text{fc} \\
        \neg b_i^\text{fc} &\rightarrow N^\text{fc}_i \leq 4, \forall i \in \{1,2\} \\ 
        \neg b_i^\text{fc} &\rightarrow Act^\text{fc}_i = \text{ReLU}, \forall i \in \{1,2\}
    \end{flalign}
\end{subequations}

\begin{table}[h]
  \caption{Table shows all hyperparameter names, types and domains of the CIFAR-NAS benchmark. The transformation column refers to post-processing computations before passing the hyperparameter value to the neural network training. }
  \label{tab:cifar_test}
  \centering
  \begin{tabular}{lllll}
    \midrule
    \# & Name & Type & Domain & Transformation \\
    \midrule
    0 & Batch size       & integer & $\left[2, 4\right]$ & $N_\text{batch} = 2^{x_0}$ \\
    1 & Learning rate    & conti. & $\left[-5.0, -1.0\right]$ & $\alpha = 10^{x_1}$ \\
    \midrule
    & \textbf{Convolutional layer 1}     & & & \\
    \midrule
    2 & Layer is active            & binary & $\{0, 1\}$ & $b_1^\text{conv} = x_2$ \\
    3 & Number of channels         & integer & $\left[2, 4\right]$ & $C^\text{conv}_1 = 2^{x_3}$ \\
    4 & Max pooling is active      & binary & $\{0, 1\}$ & $b_1^\text{pool} = x_4$ \\
    5 & Filter size                & integer & $\left[2, 5\right]$ & $F_1 = x_5$ \\
    6 & Stride                     & integer & $\left[1, 3\right]$ & $S_1 = x_6$ \\
    7 & Padding                    & integer & $\left[0, 3\right]$ & $P_1 = x_7$ \\
    8 & Activation function        & categ. & $\{\text{\footnotesize{ReLU}}, 
                                                    \text{\footnotesize{PReLU}}, 
                                                    \text{\footnotesize{Leaky ReLU}}\}$ & 
                                                    $Act^\text{conv}_1 = x_8$ \\
    \midrule
    & \textbf{Convolutional layer 2}     & & & \\
    \midrule
    9 & Layer is active            & binary & $\{0, 1\}$ & $b_2^\text{conv} = x_9$ \\
    10 & Number of channels         & integer & $\left[2, 4\right]$ & $C^\text{conv}_2 = 2^{x_{10}}$ \\
    11 & Max pooling is active      & binary & $\{0, 1\}$ & $b_2^\text{pool} = x_{11}$ \\
    12 & Filter size                & integer & $\left[2, 5\right]$ & $F_2 = x_{12}$ \\
    13 & Stride                     & integer & $\left[1, 3\right]$ & $S_2 = x_{13}$ \\
    14 & Padding                    & integer & $\left[0, 3\right]$ & $P_2 = x_{14}$ \\
    15 & Activation function        & categ. & $\{\text{\footnotesize{ReLU}}, 
                                                    \text{\footnotesize{PReLU}}, 
                                                    \text{\footnotesize{Leaky ReLU}}\}$ &  $Act^\text{conv}_2 = x_{15}$ \\
    \midrule
    & \textbf{Convolutional layer 3}     & & & \\
    \midrule
    16 & Layer is active            & binary & $\{0, 1\}$ & $b_3^\text{conv} = x_{16}$ \\
    17 & Number of channels         & integer & $\left[2, 4\right]$ & $C^\text{conv}_3 = 2^{x_{17}}$ \\
    18 & Max pooling is active      & binary & $\{0, 1\}$ & $b_3^\text{pool} = x_{18}$ \\
    19 & Filter size                & integer & $\left[2, 5\right]$ & $F_3 = x_{19}$ \\
    20 & Stride                     & integer & $\left[1, 3\right]$ & $S_3 = x_{20}$ \\
    21 & Padding                    & integer & $\left[0, 3\right]$ & $P_3 = x_{21}$ \\
    22 & Activation function        & categ. & $\{\text{\footnotesize{ReLU}}, 
                                                    \text{\footnotesize{PReLU}}, 
                                                    \text{\footnotesize{Leaky ReLU}}\}$ &  $Act^\text{conv}_3 = x_{22}$ \\
    \midrule
    & \textbf{Fully-connected layer 1}     & & & \\
    \midrule
    23 & Layer is active          & binary & $\{0, 1\}$ & $b_1^\text{fc} = x_{23}$ \\
    24 & Number of nodes          & integer & $\left[2, 7\right]$ & $N^\text{fc}_1 = 2^{x_{24}}$ \\
    25 & Activation function      & categ. & $\{\text{\footnotesize{ReLU}}, 
                                                    \text{\footnotesize{PReLU}}, 
                                                    \text{\footnotesize{Leaky ReLU}}\}$ &  $Act^\text{fc}_1 = x_{25}$ \\
    \midrule
    & \textbf{Fully-connected layer 2}     & & & \\
    \midrule
    26 & Layer is active          & binary & $\{0, 1\}$ & $b_2^\text{fc} = x_{26}$ \\
    27 & Number of nodes          & integer & $\left[2, 7\right]$ & $N^\text{fc}_2 = 2^{x_{27}}$ \\
    28 & Activation function      & categ. & $\{\text{\footnotesize{ReLU}}, 
                                                    \text{\footnotesize{PReLU}}, 
                                                    \text{\footnotesize{Leaky ReLU}}\}$ &  $Act^\text{fc}_2 = x_{28}$ \\
    \bottomrule
  \end{tabular}
\end{table}

\clearpage

\section{Additional Results}
This section presents additional results supporting the numerical evaluation in Section~\ref{sec:numerical_studies}.

\begin{figure}[ht]
    \centering
    \subfigure[Relative Model Error, Rastrigin (10D)]{\label{fig:app_unc_rastrigin_err}
        \includegraphics[height=3.8cm]{figures/unc_rastrigin_err.pdf}}
    \subfigure[Prediction Mean, Rastrigin (10D)]{\label{fig:app_unc_rastrigin_mean}
        \includegraphics[height=3.8cm]{figures/unc_rastrigin_mean.pdf}}
    \subfigure[Relative Model Error, Schwefel (10D)]{\label{fig:app_unc_schwefel_err}
        \includegraphics[height=3.8cm]{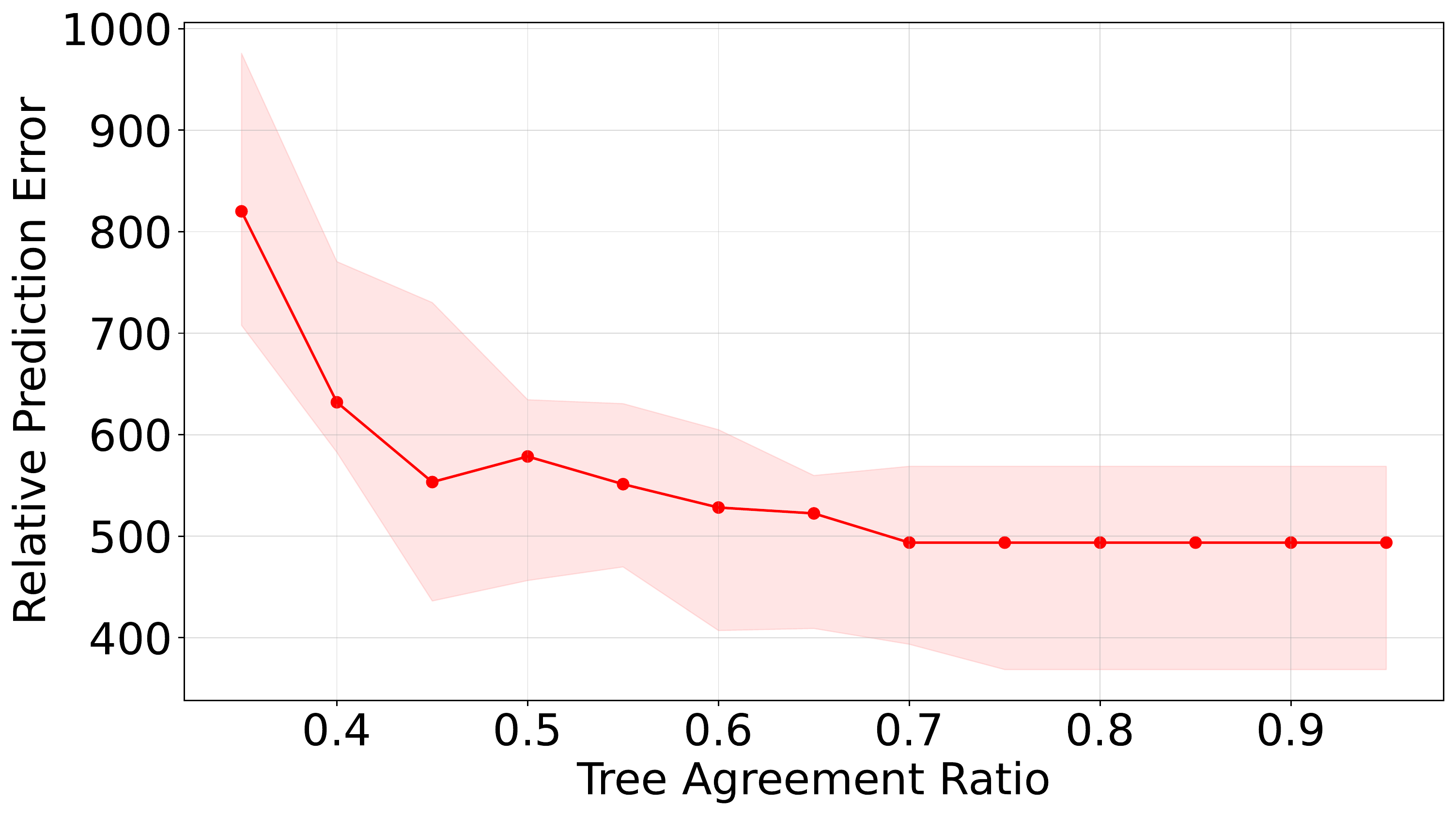}}
    \subfigure[Prediction Mean, Schwefel (10D)]{\label{fig:app_unc_schwefel_mean}
        \includegraphics[height=3.8cm]{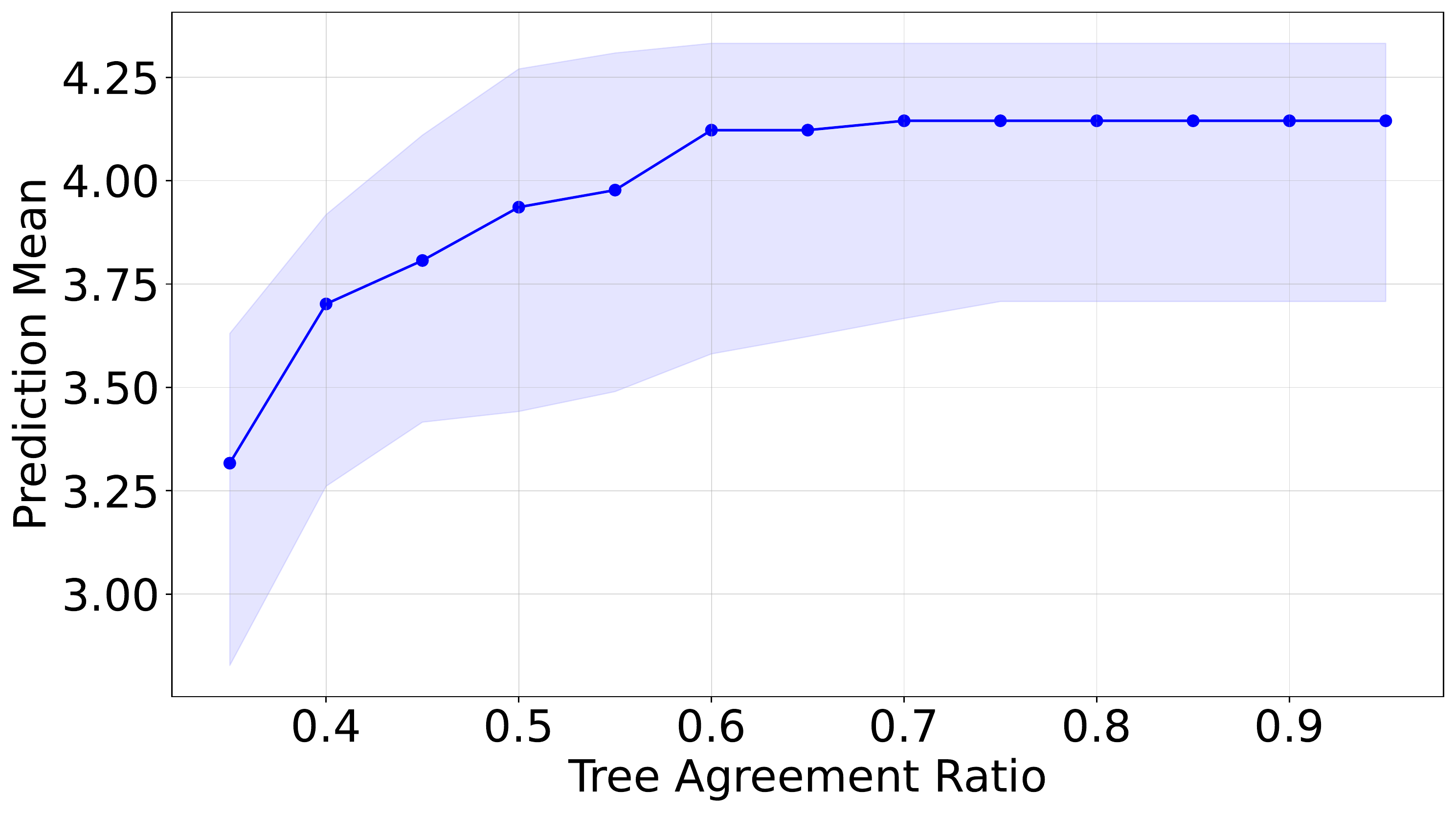}}  
    \caption{The relative prediction error (Eq.~13) and model prediction mean over the maximum tree agreement ratio $R$ for benchmark problem Schwefel (10D). Changing $R$ is equivalent to changing the maximum kernel covariance. Plot shows the median line and confidence intervals (first and third quartile) from 20 random seeds. Section~\ref{sec:uncertainty_metric} provides more details.}
    \label{fig:unc_test}
\end{figure}

\begin{figure}[ht]
    \centering
    \subfigure[Hartmann (6D)]{\label{fig:app_hartmann6d}
        \includegraphics[height=4.43cm]{figures/hartmann6d.pdf}}
    \subfigure[Rastrigin (10D)]{\label{fig:app_rastrigin}
        \includegraphics[height=4.43cm]{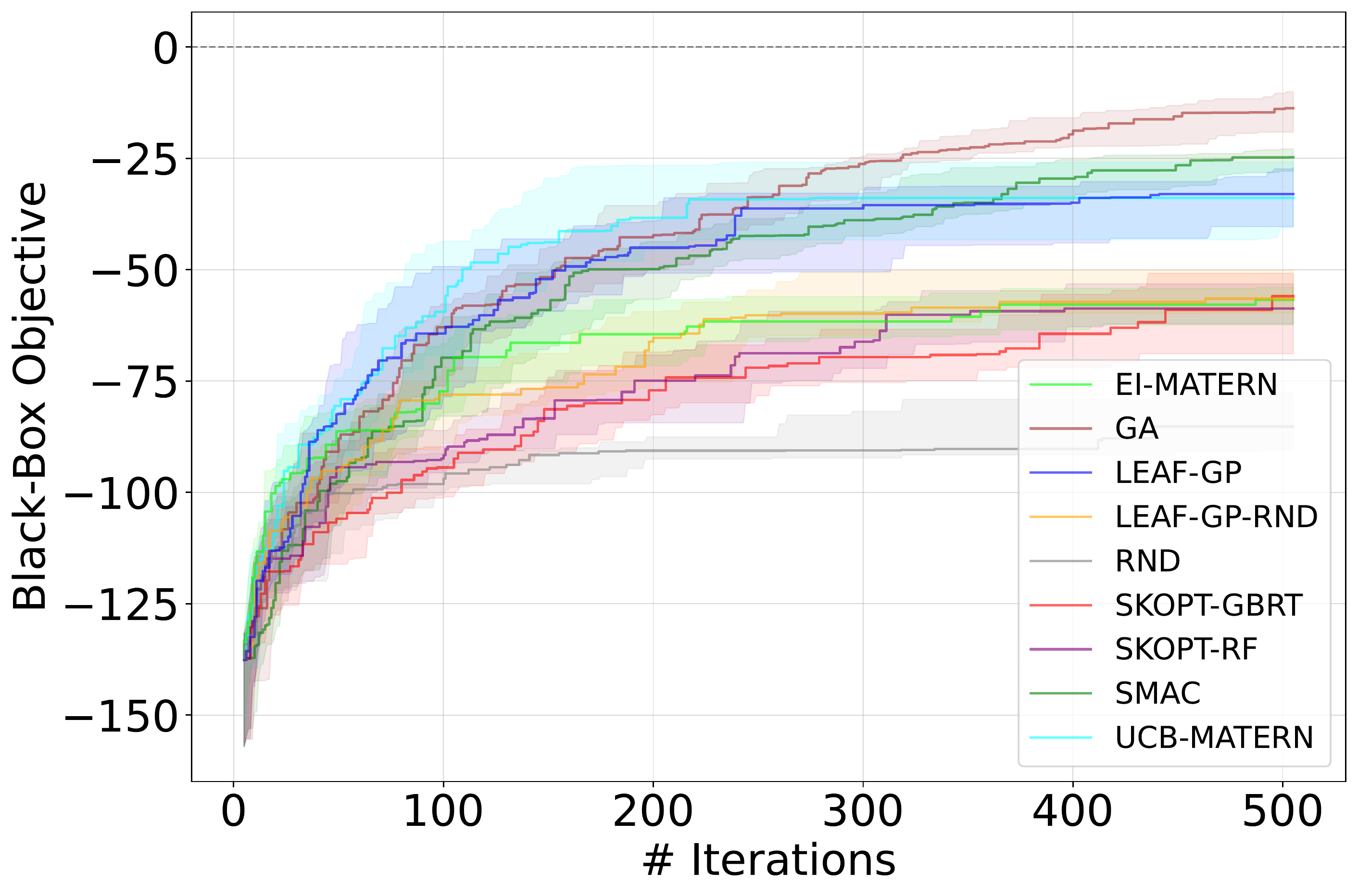}}
    \subfigure[Schwefel (10D)]{\label{fig:app_schwefel}
        \includegraphics[height=4.43cm]{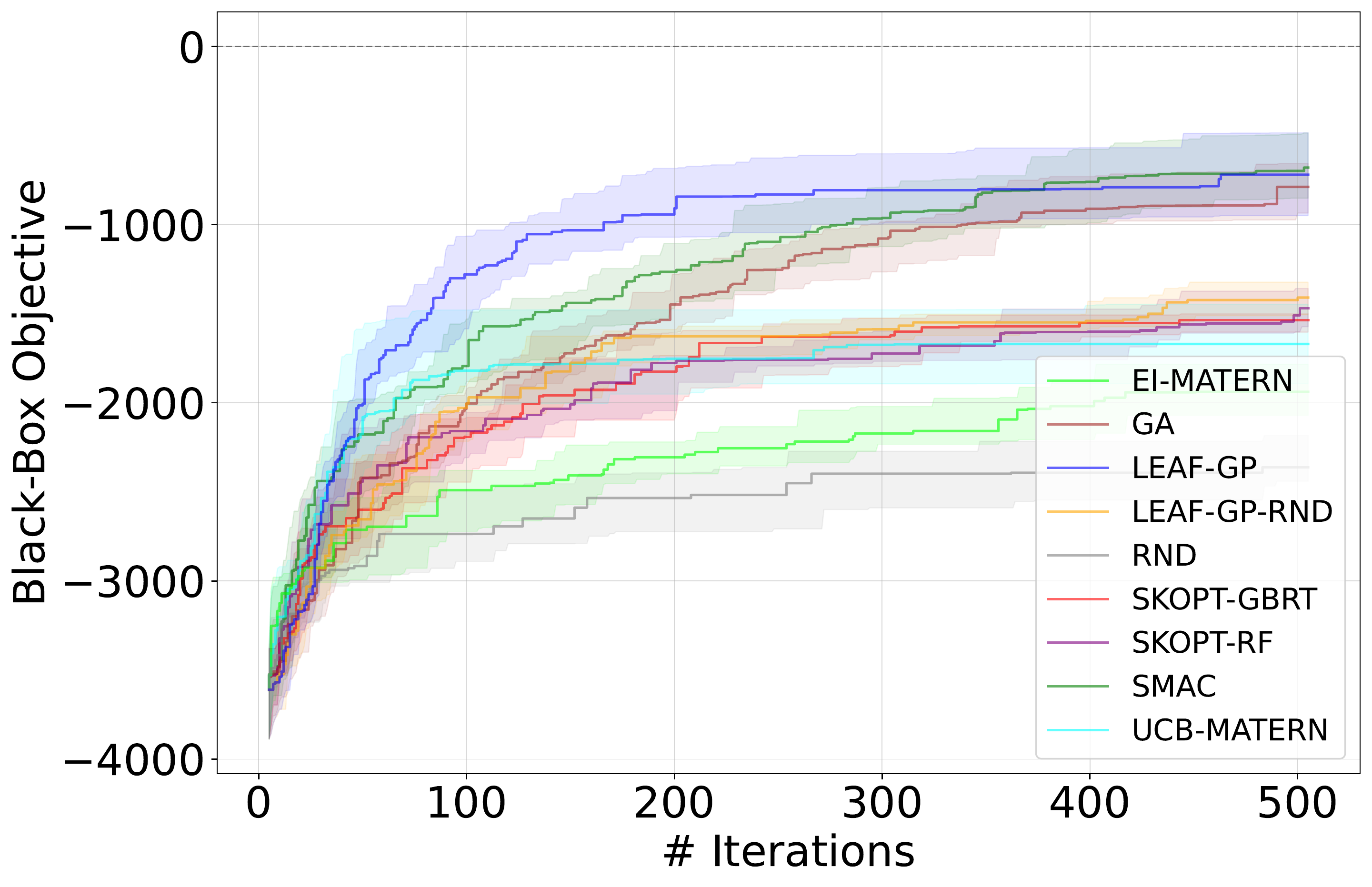}}
    \subfigure[Styblinski-Tang (10D)]{\label{fig:app_styblinski_tang}
        \includegraphics[height=4.43cm]{figures/styblinski_tang.pdf}}   
    \caption{Black-box optimization progress of \texttt{LEAF-GP} vs.\ baseline. Plot shows the median line and confidence intervals (first and third quartile) from 20 random seeds. Section~\ref{sec:local_vs_global} provides more details}
    \label{fig:app_basic_test}
\end{figure}

\clearpage

\section{VAE-NAS} \label{app:vae_nas}

\begin{figure}[ht]
    \centering
    \includegraphics[height=5.6cm]{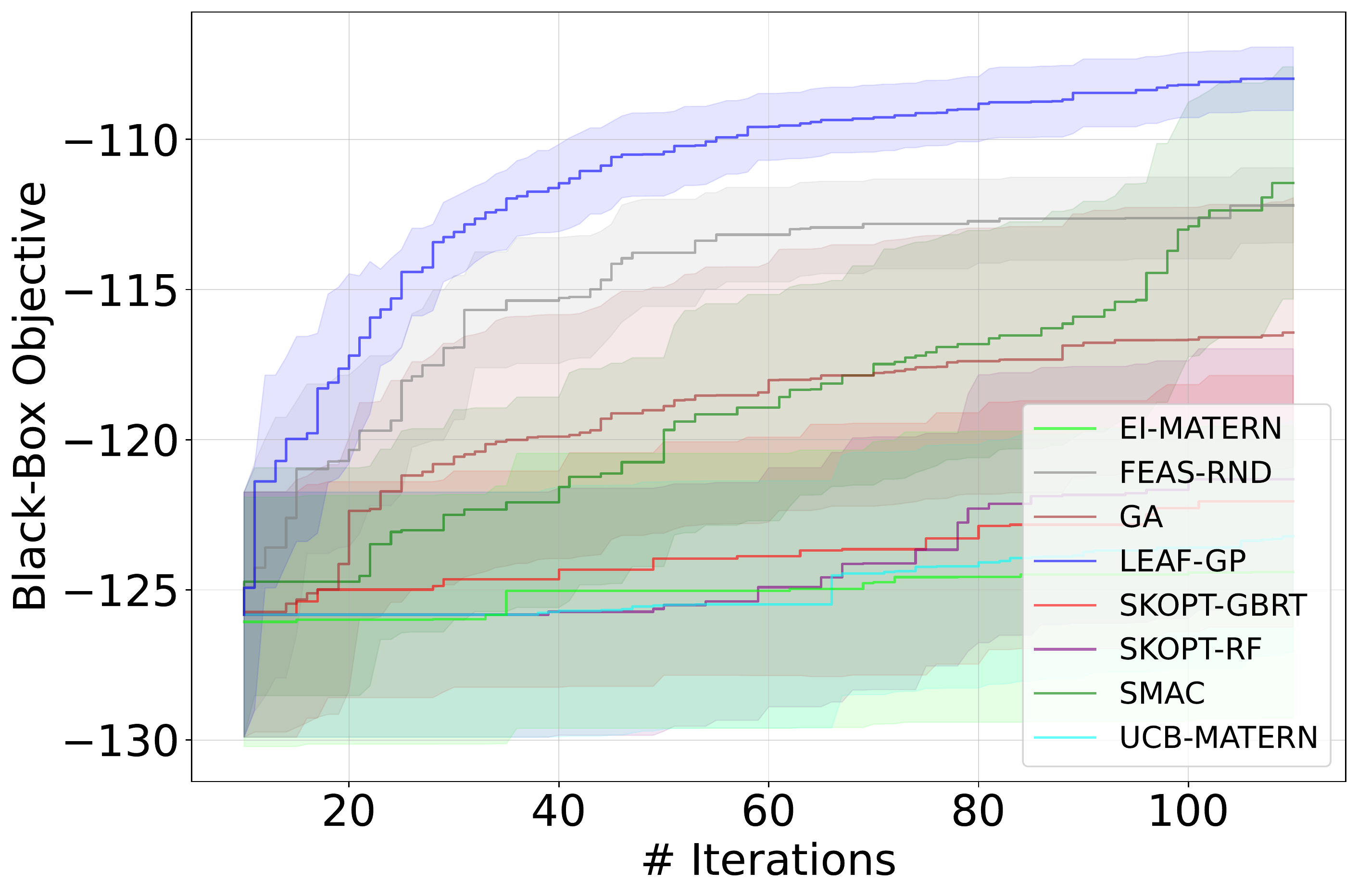}
    \caption{Feasible black-box optimization comparing \texttt{LEAF-GP} vs.\ baseline. Plot shows median line and confidence intervals (first and third quartiles) from 20 random seeds. 
    Figure subtitles give the number of dimensions (D) and inequality constraints (IC).
    Section~\ref{sec:mixed_vars} provides details.}
    \label{fig:vae_small}
\end{figure}

Table~\ref{tab:vae_test} gives more details on the Section~\ref{sec:mixed_vars} Variational Autoencoder Neural Architecture Search (VAE-NAS) benchmark problem, which was adapted from \citet{daxberger2019mixed}.
As their exact implementation is not publicly available, we created a benchmark problem based on the paper description and training scripts by \citet{vaeImplementation}.
VAE-NAS (32D) has a total of 32 hyperparameters to tune, i.e.,\ 1 continuous, 20 integer, and 11 categorical variables.
The goal is to select hyperparameter values for a variational autoencoder (VAE) \citep{kingma2013auto} trained in PyTorch \citep{NEURIPS2019_9015} on the MNIST dataset \citep{deng2012mnist} that minimize test loss, i.e.,\ the average loss of when encoding and decoding images from the test set.
We train each tested VAE for 32 epochs using the Adam solver \citep{kingma2014adam} and a fixed batch size of 128.
Only certain combinations of stride, padding, and filter size for various layers result in feasible neural architectures.
The convolutional layer output size $W^e_\text{out}$ is computed as:
\begin{equation}
    W^e_\text{out} = \frac{W^e_\text{in}-F^\text{e}+2P^\text{e}}{S^\text{e}} + 1 \in \mathbb{N},
\end{equation}
Decoder constraints are more complicated given that the size of the VAE output must match the original MNIST image size.
The output of deconvolutional layers is computed according to:
\begin{equation}
\label{eq:deconv_layer}
    W^d_\text{out} = S^\text{d}(W^d_\text{in} - 1) + F^\text{d}-2P^\text{d} + O^\text{d}
\end{equation}
Matching the output and input sizes of the VAE is non-trivial, as hyperparameters of the convolutional and deconvolutional layers are themselves set by the optimization algorithms. 
To simplify the training, layer inputs are parameterized based on outputs of the previous layer for convolutional, deconvolutional, and fully-connected layers.
Specifically, fully-connected layers \texttt{FC2} and \texttt{FC3} have $2 \times N^\text{lat}$ nodes.
According to \citet{daxberger2019mixed}, we parameterize the size of the last fully-connected layer---which can be \texttt{FC3}, \texttt{FC4}, or the latent space layer depending on which layers are active---as $C^\text{d}_1 \times 7 \times 7$.
This allows methods not supporting explicit input constraints to still easily find a feasible neural architecture by deactivating all deconvolutional layers and computing an output $16 \times 7 \times 7 = 784$ with $C^\text{d}_1 = 16$ (the original MNIST image size is $1 \times 28 \times 28 = 784$).
This rule supersedes other layer size definitions for fully-connected layers. 
Moreover, we extend the benchmark by introducing the nonlinear activation function of each layer as a categorical optimization hyperparameter.
Methods that do not support categorical features use one-hot encoding.
The activation function of the last layer is fixed as the sigmoid function, superseding other activation function hyperparameters.
The VAE is trained using the sum of binary cross-entropy loss (reconstruction error) and KL divergence.
We use the same loss function to evaluate the VAE's performance on the test dataset, giving the black-box objective in Fig.~\ref{fig:vae_small}.
To allow for a fair comparison, the same 10 randomly sampled feasible architectures initialize all methods.

\texttt{LEAF-GP} has access to constraints describing feasible neural architectures. We begin by defining auxiliary variables similar to the benchmark in Section~\ref{app:cifar_nas}:
\begin{subequations}
    \label{eq:enc_con_gen}
    \begin{flalign}
        W^\text{e}_{\text{in},i} &\in \mathbb{N}_0, &\forall i \in \left[1,2\right], \\
        w^\text{e}_{\text{out},i} &\in \mathbb{N}_0, &\forall i \in \left[1,2\right], \\
        W^\text{e}_{\text{out},i} &\in \mathbb{N}_0, &\forall i \in \left[1,2\right], \\
        b^\text{e}_{\text{covn},i} &\in \{0,1\}, &\forall i \in \left[1,2\right], \\
        N^\text{e}_\text{conv} &= b^\text{e}_{\text{conv},1} + b^\text{e}_{\text{conv},2}, & \label{eq:enc_con_gen_e}\\
        b^\text{e}_{\text{conv},1} &\geq b^\text{e}_{\text{conv},2} \label{eq:enc_con_gen_f}&
    \end{flalign}
\end{subequations}
where $W^\text{e}_{\text{in},i}$ and $W^\text{e}_{\text{out},i}$ denote, respectively, the input and output sizes of convolutional layer $i$ in the encoder.
We also define an auxiliary variable $w^\text{e}_{\text{out},i}$ to track the output size for inactive layers, as well as binary variables $b^\text{e}_{\text{conv},i}$ corresponding to the active/inactive state of each layer.  
Eq.~\eqref{eq:enc_con_gen_e} and Eq.~\eqref{eq:enc_con_gen_f} link binary variables $b^\text{e}_{\text{conv},i}$ to the number of active convolutional layers in the encoder.
Using these auxiliary variables, the following relations can be expressed: 
\begin{subequations}
    \label{eq:enc_con}
    \begin{flalign}
    W^\text{e}_{\text{in},1} &= 28, & \label{eq:enc_con_a}\\
    W^\text{e}_{\text{in},2} &=  W^\text{e}_{\text{out},1}, & \label{eq:enc_con_b}\\
    w^\text{e}_{\text{out},i} &= \frac{W^\text{e}_{\text{in},i} - F^\text{e}_i + 2P^\text{e}_i}{S^\text{e}_i} + 1, &\forall i \in \left[1,2\right], \label{eq:enc_con_c}\\
    W^\text{e}_{\text{out},i} &= b^\text{e}_{\text{conv},i} w^\text{e}_{\text{out},i} + (1 - b^\text{e}_{\text{conv},i}) W^\text{e}_{\text{in},i}, &\forall i \in \left[1,2\right], \label{eq:enc_con_d}\\
    W^\text{e}_{\text{out},2} &\geq 1 & \label{eq:enc_con_e}
    \end{flalign}
\end{subequations}
Eq.~\eqref{eq:enc_con_a} and Eq.~\eqref{eq:enc_con_b} define the input sizes as the MNIST image input size  $W^\text{e}_{\text{in},1} = 28$ for the first layer and the output size of the previous convolution for ensuing layers. 
Eq.~\eqref{eq:enc_con_c} defines the layer $i$ output $w^\text{e}_{\text{out},i}$ given the filter size $F^\text{e}_i$, padding $P^\text{e}_i$, and stride $S^\text{e}_i$.
Eq.~\eqref{eq:enc_con_d} ensures that the actual convolutional layer output $W^\text{e}_{\text{out},i}$ only takes the value of $w^\text{e}_{\text{out},i}$ if the layer is active.
Finally, Eq.~\eqref{eq:enc_con_e} enforces the output size of the encoder to be at least one.

We add similar auxiliary variables and constraints for the deconvolutional layers:
\begin{subequations}
    \label{eq:dec_deconv_gen}
    \begin{flalign}
        W^\text{d}_{\text{in},i} &\in \mathbb{N}_0, &\forall i \in \left[1,2\right], \\
        w^\text{d}_{\text{out},i} &\in \mathbb{N}_0, &\forall i \in \left[1,2\right], \\
        W^\text{d}_{\text{out},i} &\in \mathbb{N}_0, &\forall i \in \left[1,2\right], \\
        b^\text{d}_{\text{dec},i} &\in \{0,1\}, \forall i \in \left[1,2\right], \\
        N^\text{d}_\text{dec} &= b^\text{e}_{\text{dec},1} + b^\text{e}_{\text{dec},2}, & \\
        b^\text{d}_{\text{dec},1} &\geq b^\text{e}_{\text{dec},2}, & \\
        b^\text{d}_{\text{dec},1} &\rightarrow W^\text{d}_{\text{out},2} = 28, & \label{eq:dec_deconv_gen_g}\\
        \neg b^\text{d}_{\text{dec},1} &\rightarrow C^\text{d}_1 = 16 & \label{eq:dec_deconv_gen_h}
    \end{flalign}
\end{subequations}
The Eq.~\eqref{eq:dec_deconv_gen_g} indicator constraint restricts the decoder output to be the original image size $W^\text{d}_{\text{out},2} = 28$ if deconvolutional layers are active.
Another indicator constraint Eq.~\eqref{eq:dec_deconv_gen_h} handles the aforementioned case where no deconvolutional layer is active and $C^\text{d}_1 = 16$ ensures that the decoder output size can be resized to original MNIST image size.
We emphasize that this rule is introduced to simplify the feasible architecture search for methods that do not support explicit input constraints.
Note that \texttt{LEAF-GP} could add additional constraints to ensure the architecture's output size can always be resized to the original image size of 28$\times$28.
\begin{subequations}
    \label{eq:enc_deconv}
    \begin{flalign}
    W^\text{d}_{\text{in},1} &= 7, &\\
    W^\text{d}_{\text{in},2} &= W^\text{d}_{\text{out},1}, &\\
    w^\text{d}_{\text{out},i} &= S^\text{d}_i(W^\text{d}_{\text{in},i} - 1) + F^\text{d}_i - 2P^\text{d}_i + O^\text{d}_i, &\forall i \in \left[1,2\right],\\
    S^\text{d}_i &\geq O^\text{d}_i + 1, &\forall i \in \left[1,2\right], \label{eq:enc_deconv_d}\\
    W^\text{d}_{\text{out},i} &= b^\text{d}_{\text{conv},i} w^\text{d}_{\text{out},i} + (1 - b^\text{d}_{\text{conv},i}) W^\text{d}_{\text{in},i}, &\forall i \in \left[1,2\right]
    \end{flalign}
\end{subequations}
Similar to the encoder, Eq.~\eqref{eq:enc_deconv} defines constraints for feasible decoder layers.
For deconvolutional layers, we also tune output padding $O^\text{d}_i$.
According to the PyTorch \citep{gardner2018gpytorch} documentation, output padding must be smaller than either stride or dilation.
Given that we do not optimize dilation in deconvolutional layers, Eq.~\eqref{eq:enc_deconv_d} enforces output padding to be smaller than stride.
We introduce similar constraints for fully-connected layers in both the encoder and decoder:
\begin{subequations}
    \label{eq:enc_gen_feas}
    \begin{flalign}
        b_{\text{fc},i} &\in \{0,1\}, &\forall i \in \left[1,4\right], \\
        N_\text{fc} &= b_{\text{fc},1} + b_{\text{fc},2}, & \\
        N^\text{d}_\text{fc} &= b_{\text{fc},3} + b_{\text{fc},4}, & \\
        b_{\text{fc},1} &\geq b_{\text{fc},2}, & \\
        b_{\text{fc},3} &\geq b_{\text{fc},4} &
    \end{flalign}
\end{subequations}
To break symmetries in the benchmark problem, we add constraints~\eqref{eq:hier_vae_a}--\eqref{eq:hier_vae_m}:
\begin{subequations}
    \label{eq:hier_vae}
    \begin{flalign}
        \neg b^\text{e}_{\text{conv},i} &\rightarrow C^\text{e}_i \leq 4, &\forall i \in \left[1,2\right], \label{eq:hier_vae_a}\\
        \neg b^\text{e}_{\text{conv},i} &\rightarrow S^\text{e}_i \leq 1, &\forall i \in \left[1,2\right], \\
        \neg b^\text{e}_{\text{conv},i} &\rightarrow P^\text{e}_i \leq 0, &\forall i \in \left[1,2\right], \\
        \neg b^\text{e}_{\text{conv},i} &\rightarrow F^\text{e}_i = 2, &\forall i \in \left[1,2\right], \\
        \neg b^\text{e}_{\text{conv},i} &\rightarrow Act^\text{e}_i = \text{ReLU}, &\forall i \in \left[1,2\right], \\
        \neg b_{\text{fc},i} &\rightarrow Act^\text{fc}_i = \text{ReLU}, &\forall i \in \left[1,4\right], \\
        \neg b_{\text{fc},1} &\rightarrow N^\text{fc}_1 \leq 0, & \\
        \neg b^\text{d}_{\text{dec},i} &\rightarrow C^\text{d}_i \leq 4, &\forall i \in \left[1,2\right], \\
        \neg b^\text{d}_{\text{dec},i} &\rightarrow S^\text{d}_i \leq 1, &\forall i \in \left[1,2\right], \\
        \neg b^\text{d}_{\text{dec},i} &\rightarrow P^\text{d}_i \leq 0, &\forall i \in \left[1,2\right], \\
        \neg b^\text{d}_{\text{dec},i} &\rightarrow O^\text{d}_i \leq 0, &\forall i \in \left[1,2\right], \\
        \neg b^\text{d}_{\text{dec},i} &\rightarrow F^\text{d}_i = 2, &\forall i \in \left[1,2\right], \\
        \neg b^\text{d}_{\text{dec},1} &\rightarrow Act^\text{d}_1 = \text{ReLU} & \label{eq:hier_vae_m}
    \end{flalign}
\end{subequations}
Constraints~\eqref{eq:hier_vae_a}--\eqref{eq:hier_vae_m} set layer-specific hyperparameters to pre-defined default values when the associated layer is inactive.
We select these defaults as the lower bound for non-categorical variables and the first category for categorical variables.
While \texttt{SMAC} is unable to handle more complicated constraints restricting outputs of deconvolutional layers, it can handle hierarchical search space structures.
For VAE-NAS benchmark runs using \texttt{SMAC} as an optimizer we enforce hierarchies according to constraints~\eqref{eq:hier_vae_a}--\eqref{eq:hier_vae_m} which deactivate hyperparameters for inactive layers.

\begin{table}[h]
  \caption{Hyperparameter names, types, and domains for the VAE-NAS benchmark. The transformation column refers to post-processing computations before passing the hyperparameter value to the neural network training. The architecture with all layers activated comprises  \texttt{C1-C2-FC1-FC2-L-FC3-FC4-D1-D2}, with \texttt{L} referring to the latent space layer.}
  \label{tab:vae_test}
  \centering
  \begin{tabular}{lllll}
    \midrule
    \# & Name & Type & Domain & Transformation \\
    \midrule
    \midrule
    & \large \textbf{General}     & & & \\
    \midrule
    0 & \; \; Learning rate                     & conti. & $\left[-4.0, -2.0\right]$ & $\alpha = 10^{x_0}$ \\
    1 & \; \; Latent space size                 & integer & $\left[16, 64\right]$ & $N^\text{lat} = x_1$ \\
    2 & \; \; Num. conv. enc. layers    & integer & $\left[0, 2\right]$ & $N^\text{e}_\text{conv} = x_2$ \\
    3 & \; \; Num. fully-conn. enc. layers      & integer & $\left[0, 2\right]$ & $N^\text{e}_\text{fc} = x_3$ \\
    4 & \; \; Num. deconv. dec. layers  & integer & $\left[0, 2\right]$ & $N^\text{d}_\text{dec} = x_4$ \\
    5 & \; \; Num. fully-conn. dec. layers      & integer & $\left[0, 2\right]$ & $N^\text{d}_\text{fc} = x_5$ \\
    \midrule
    \midrule
    & \large \textbf{Encoder}     & & & \\
    \midrule
    & \; \textbf{Convolutional layer 1} (\texttt{C1})    & & & \\
    6 & \; \; Number of output channels  & integer & $\left[2, 5\right]$ & $C^\text{e}_1 = 2^{x_6}$ \\
    7 & \; \; Stride                     & integer & $\left[1, 2\right]$ & $S^\text{e}_1 = x_7$ \\
    8 & \; \; Padding                    & integer & $\left[0, 3\right]$ & $P^\text{e}_1 = x_8$ \\
    9 & \; \; Filter size                & categ. & $\{3, 5\}$ & $F^\text{e}_1 = x_9$ \\
    10 & \; \; Activation function        & categ. & $\{\text{\footnotesize{ReLU}}, 
                                                    \text{\footnotesize{PReLU}}, 
                                                    \text{\footnotesize{Leaky ReLU}}\}$ & 
                                                    $Act^\text{e}_1 = x_{10}$ \\
    \midrule
    & \; \textbf{Convolutional layer 2} (\texttt{C2})     & & & \\
    11 & \; \; Number of output channels  & integer & $\left[3, 6\right]$ & $C^\text{e}_2 = 2^{x_{11}}$ \\
    12 & \; \; Stride                     & integer & $\left[1, 2\right]$ & $S^\text{e}_2 = x_{12}$ \\
    13 & \; \; Padding                    & integer & $\left[0, 3\right]$ & $P^\text{e}_2 = x_{13}$ \\
    14 & \; \; Filter size                & categ. & $\{3, 5\}$ & $F^\text{e}_2 = x_{14}$ \\
    15 & \; \; Activation function        & categ. & $\{\text{\footnotesize{ReLU}}, 
                                                    \text{\footnotesize{PReLU}}, 
                                                    \text{\footnotesize{Leaky ReLU}}\}$ & 
                                                    $Act^\text{e}_2 = x_{15}$ \\
    \midrule
    & \; \textbf{Fully-connected layer 1} (\texttt{FC1})     & & & \\
    16 & \; \; Number of nodes          & integer & $\left[0, 15\right]$ & $N^\text{fc}_1 = 64 \times x_{16}$ \\
    17 & \; \; Activation function      & categ. & $\{\text{\footnotesize{ReLU}}, 
                                                    \text{\footnotesize{PReLU}}, 
                                                    \text{\footnotesize{Leaky ReLU}}\}$ &  $Act^\text{fc}_1 = x_{17}$ \\
    \midrule
    & \; \textbf{Fully-connected layer 2} (\texttt{FC2})     & & & \\
    18 & \; \; Activation function      & categ. & $\{\text{\footnotesize{ReLU}}, 
                                                    \text{\footnotesize{PReLU}}, 
                                                    \text{\footnotesize{Leaky ReLU}}\}$ &  $Act^\text{fc}_2 = x_{18}$ \\
    \midrule
    \midrule
    & \large \textbf{Decoder}     & & & \\
    \midrule
    & \; \textbf{Fully-connected layer 3} (\texttt{FC3})    & & & \\
    19 & \; \; Activation function      & categ. & $\{\text{\footnotesize{ReLU}}, 
                                                    \text{\footnotesize{PReLU}}, 
                                                    \text{\footnotesize{Leaky ReLU}}\}$ &  $Act^\text{fc}_3 = x_{19}$ \\
    \midrule
    & \; \textbf{Fully-connected layer 4} (\texttt{FC4})     & & & \\
    20 & \; \; Activation function      & categ. & $\{\text{\footnotesize{ReLU}}, 
                                                    \text{\footnotesize{PReLU}}, 
                                                    \text{\footnotesize{Leaky ReLU}}\}$ &  $Act^\text{fc}_4 = x_{20}$ \\
    \midrule
    & \; \textbf{Deconvolutional layer 1} (\texttt{D1})     & & & \\
    21 & \; \; Number of input channels   & integer & $\left[3, 6\right]$ & $C^\text{d}_1 = 2^{x_{21}}$ \\
    22 & \; \; Stride                     & integer & $\left[1, 2\right]$ & $S^\text{d}_1 = x_{22}$ \\
    23 & \; \; Padding                    & integer & $\left[0, 3\right]$ & $P^\text{d}_1 = x_{23}$ \\
    24 & \; \; Output Padding             & integer & $\left[0, 1\right]$ & $O^\text{d}_1 = x_{24}$ \\
    25 & \; \; Filter size                & categ. & $\{3, 5\}$ & $F^\text{d}_1 = x_{25}$ \\
    26 & \; \; Activation function        & categ. & $\{\text{\footnotesize{ReLU}}, 
                                                    \text{\footnotesize{PReLU}}, 
                                                    \text{\footnotesize{Leaky ReLU}}\}$ & 
                                                    $Act^\text{d}_1 = x_{26}$ \\
    \midrule
    & \; \textbf{Deconvolutional layer 2} (\texttt{D2})    & & & \\
    27 & \; \; Number of input channels   & integer & $\left[2, 5\right]$ & $C^\text{d}_2 = 2^{x_{27}}$ \\
    28 & \; \; Stride                     & integer & $\left[1, 2\right]$ & $S^\text{d}_2 = x_{28}$ \\
    29 & \; \; Padding                    & integer & $\left[0, 3\right]$ & $P^\text{d}_2 = x_{29}$ \\
    30 & \; \; Output Padding             & integer & $\left[0, 1\right]$ & $O^\text{d}_2 = x_{30}$ \\
    31 & \; \; Filter size                & categ. & $\{3, 5\}$ & $F^\text{d}_2 = x_{31}$ \\
    \bottomrule
  \end{tabular}
\end{table}


\end{document}